\pdfoutput=1
\documentclass[11pt, letterpaper, shortlabels]{berkeley}

\usepackage{hyperref}
\usepackage{color-edits}
\usepackage{algorithm}
\usepackage{algorithmicx}
\usepackage{algpseudocode}
\usepackage{soul}

\ifx\assumption\undefined
\newtheorem{assumption}{Assumption}
\fi

\usepackage[capitalize,noabbrev]{cleveref}
\makeatletter
\def\adl@drawiv#1#2#3{%
        \hskip.5\tabcolsep
        \xleaders#3{#2.5\@tempdimb #1{1}#2.5\@tempdimb}%
                #2\z@ plus1fil minus1fil\relax
        \hskip.5\tabcolsep}
\newcommand{\cdashlinelr}[1]{%
  \noalign{\vskip\aboverulesep
           \global\let\@dashdrawstore\adl@draw
           \global\let\adl@draw\adl@drawiv}
  \cdashline{#1}
  \noalign{\global\let\adl@draw\@dashdrawstore
           \vskip\belowrulesep}}
\makeatother

\usepackage{wrapfig}
\newcommand{\rmnum}[1]{\romannumeral #1}
\newcommand{\Rmnum}[1]{\uppercase\expandafter{\romannumeral #1}}

\title{AutoBench-V: Can Large Vision-Language Models Benchmark Themselves?}

\usepackage[all]{hypcap}
\usepackage{xcolor}

\usepackage[authoryear, round]{natbib}

\usepackage{hyperref}[citecolor=gold,linkcolor=gold]

\hypersetup{
    colorlinks = true,
    citecolor = {gold},
}

\usepackage{multirow, makecell, caption}\usepackage{microtype}
\usepackage{graphicx}
\usepackage{booktabs} 

\usepackage{amsmath}
\usepackage{amssymb}
\usepackage{mathtools}
\usepackage{amsthm}
\usepackage{mathrsfs}
\usepackage{nicefrac}
\usepackage{dsfont}
\usepackage{enumitem}
\usepackage{arydshln}
\usepackage{xspace}
\usepackage[capitalize,noabbrev]{cleveref}
\usepackage{subcaption}
\usepackage{lipsum}
\usepackage{listings}

\usepackage{amsmath}
\usepackage{amssymb}
\usepackage{mathtools}
\usepackage{amsthm}
\usepackage{bbm}
\usepackage[most]{tcolorbox}
\usepackage{fontawesome}
\usepackage{algpseudocode}
\usepackage{setspace}

\usepackage{color}
\definecolor{deepblue}{rgb}{0,0,0.5}
\definecolor{deepred}{rgb}{0.6,0,0}
\definecolor{deepgreen}{rgb}{0,0.5,0}

\newcommand\pythonstyle{\lstset{
basicstyle=\ttfamily\footnotesize,
language=Python,
morekeywords={self, clip, exp, mse_loss, uniform_sample, concatenate, logsumexp},              
keywordstyle=\color{deepblue},
emph={MyClass,__init__},          
emphstyle=\color{deepred},    
stringstyle=\color{deepgreen},
frame=single,                         
showstringspaces=false
}}

\tcbset{
    prompt/.style={
        colback=violet!5!white,         
        colframe=violet!75!black,       
        boxsep=4pt,                   
        arc=5pt,                      
        boxrule=0.8pt,                
        width=\linewidth,
        enhanced,
        drop shadow={violet!30!black},  
        boxed title style={
            colback=violet!20!white,    
            colframe=violet!50!black,   
            boxrule=0.8pt,            
            rounded corners=south,    
            size=small,
            boxsep=3pt,               
            left=3mm,
            right=3mm,
            fontupper=\color{violet!30!black}\bfseries  
        },
        attach boxed title to top center={yshift=-\tcboxedtitleheight/2},  
        top=3mm,
        bottom=3mm,
        coltitle=violet!40!black        
    }
}

\lstnewenvironment{python}[1][]
{
\pythonstyle
\lstset{#1}
}
{}

\newcommand\pythoninline[1]{{\pythonstyle\lstinline!#1!}}

\makeatletter
\def\mathcolor#1#{\@mathcolor{#1}}
\def\@mathcolor#1#2#3{%
  \protect\leavevmode
  \begingroup
    \color#1{#2}#3%
  \endgroup
}
\makeatother

\Crefformat{equation}{#2Eq.\;(#1)#3}

\Crefformat{figure}{#2Figure #1#3}
\Crefformat{assumption}{#2Assumption #1#3}
\Crefname{assumption}{Assumption}{Assumptions}

\usepackage{crossreftools}
\usepackage{dsfont}
\usepackage{nicefrac}

\author[1]{Han Bao*}
\author[1]{Yue Huang*}
\author[2]{Yanbo Wang$\ddagger$}
\author[2]{Jiayi Ye$\ddagger$}
\author[2]{Xiangqi Wang}
\author[2]{Xiuying Chen}
\author[3]{Yue Zhao}
\author[4]{Tianyi Zhou}
\author[5]{Mohamed Elhoseiny}
\author[2]{Xiangliang Zhang}

\affil[1]{University of Notre Dame}
\affil[2]{MBZUAI}
\affil[3]{University of Southern California}
\affil[4]{University of Maryland, College Park}
\affil[5]{KAUST}
\affil[$\ddagger$]{Visiting students}
\affil[*]{Equal contributions}

\correspondingauthor{xzhang33@nd.edu (Xiangliang Zhang)}

\begin{abstract}
\textbf{Abstract:} Large Vision-Language Models (LVLMs) have become essential for advancing the integration of visual and linguistic information. However, the evaluation of LVLMs presents significant challenges as the evaluation benchmark always demands lots of human cost for its construction, and remains static, lacking flexibility once constructed. Even though automatic evaluation has been explored in textual modality, the visual modality remains under-explored. As a result, in this work, we address a question: ``Can LVLMs themselves be
used to benchmark each other in the visual automatically
domain?". We introduce \textsc{AutoBench-V}, an automated framework for serving evaluation on demand, \emph{i.e.}, benchmarking LVLMs based on specific aspects of model capability. \textsc{AutoBench-V} leverages text-to-image models to generate relevant image samples and then utilizes LVLMs to orchestrate visual question-answering (VQA) tasks, completing the evaluation process efficiently and flexibly. Through an extensive evaluation of nine popular LVLMs across five demanded user inputs (\emph{i.e.}, evaluation capabilities), the framework shows effectiveness and reliability. 
\end{abstract}

\begin{document}

\maketitle

\section{Introduction}

\begin{figure}[h]
\centering
    \includegraphics[width=0.9\textwidth]{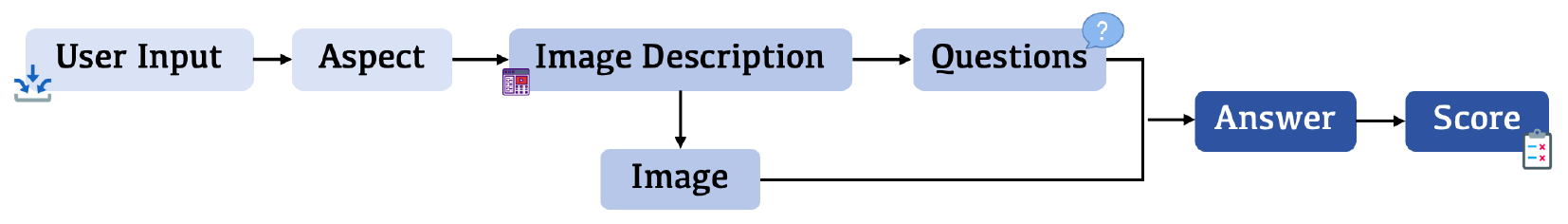}
    \captionsetup{justification=centerlast}
    \caption{An overview of the \textsc{AutoBench-V} pipeline.}
    \label{fig:automated process}
\end{figure}

In recent years, the rapid development of natural language processing (NLP) has given rise to Large Language Models (LLMs) \citep{zhao2023survey}, which have revolutionized the field and opened new avenues for innovation and practical downstream applications \citep{liu2023agentbench, liu2023deid}. As the capabilities of LLMs grow, researchers begin to explore the integration of visual information understanding capabilities into LLMs, leading to the emergence of Large Vision-Language models (LVLMs) \citep{liu2023visualinstructiontuning}. These models are trained on extensive multi-modal datasets (e.g., image-text data), enabling them to perform sophisticated cross-modal perception, understanding, and reasoning by effectively integrating visual and textual information \citep{antol2015vqa,zou2023objectdetection20years,Ghandi_2023,chen2024gui}. 

With LVLMs becoming more widely adopted, their evaluation has become increasingly important for determining their limitations and reliability.  Recent research \citep{xu2023lvlmehubcomprehensiveevaluationbenchmark,liu2023mmbenchmultimodalmodelallaround, li2023seedbenchbenchmarkingmultimodalllms, li2023seedbench2benchmarkingmultimodallarge, yin2024lamm,ying2024mmt} highlights the need for evaluation standards that cover a broad spectrum of multimodal capabilities. Although these benchmarks examine selected dimensions of LVLM performance, they lack flexibility for on-demand assessments in different capability areas. Recent studies have explored using generative models in automating evaluation, which offers flexibility in varying evaluation dimensions and reduces the human cost of benchmark dataset construction \citep{wu2024unigenunifiedframeworktextual, zhu2023dyval, li2024autobenchercreatingsalientnovel}. 
However, most of these efforts concentrate on LLMs operating solely on text, leaving the visual aspect under-explored. Motivated by this gap, we ask the following question: \textbf{\textit{“Can LVLMs themselves be used to automatically benchmark each other in the visual domain?”}} In other words, is it feasible to leverage LVLMs as both question-generators and evaluators, with minimal human intervention, to create flexible and reliable tests of multimodal understanding?

Automating LVLM evaluation presents several key challenges. 
First, the targeted capabilities to be evaluated must be identified based on the input demand. This foundation enables the generation of relevant images and appropriate VQA tasks to accurately assess the LVLMs' performance in those specific aspects\footnote{\textit{An aspect example would be like ``Spatial Understanding'' \cite{li2023seedbenchbenchmarkingmultimodalllms}}.}.
Second, the generated content must be relevant and controllable, so that it accurately tests the targeted skills. Third, there is a risk of “answer leakage”: 1) the issue arises when the model responsible for generating questions exhibits self-enhancement bias \citep{ zheng2023judging,ye2024llmjudgebias}—it can achieve higher performance as it embeds too many ``hints'' in the generated questions, 2) the prior work \citep{chen2024we} shows that models can achieve high VQA performance using text alone. 

To address the above challenges, we propose \textsc{AutoBench-V}, which supports automated LVLM evaluation based on user demand for specific evaluation aspects of model capability (\emph{e.g.}, \textit{ spatial understanding}). Initially, the input demand (i.e., evaluation aspect) is processed by an \textit{examiner LVLM}. Each aspect is further divided into several fine-grained components, for which image descriptions of varying difficulty levels are generated. To ensure that descriptions align with their corresponding images, a self-validation mechanism is applied through VQA. Furthermore, an error control mechanism is implemented to prevent a negative impact of misalignment between text and image. To avoid the potential self-enhancement bias and the situation that LVLMs only utilize textual information to guess answers \cite{chen2024we}, we introduce an error-driven option adjustment strategy to reduce such bias and ``force'' model to utilize visual information for answering. Moreover, we also introduce multi-examiners to avoid \textit{being the player and the referee at the same time}. Finally, the generated questions and images are then presented to the evaluated LVLM to generate responses, which are assessed by reference answers \citep{liu2023alignbench}. The high-level pipeline of \textsc{AutoBench-V} is shown in \autoref{fig:automated process}.

\begin{figure}[t]
    \centering
    \includegraphics[width=1.0\textwidth]{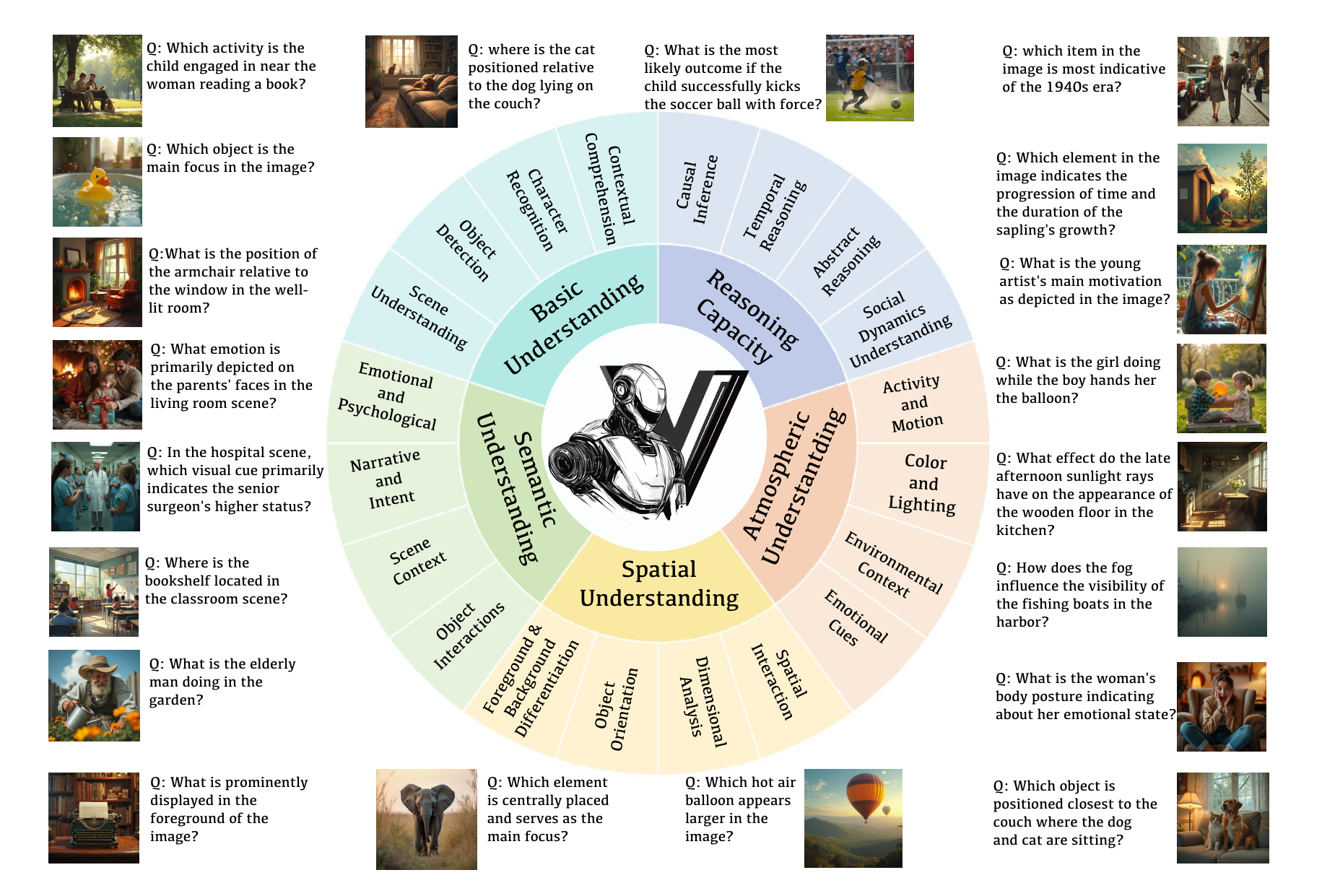}
    \caption{Five key evaluation dimensions in \textsc{AutoBench-V}—Basic Understanding, Reasoning Capacity, Semantic Understanding, Spatial Understanding, and Atmospheric Understanding—and their sub-aspects, illustrated by the representative question–image pairs.}
    \label{fig:user input}
\end{figure}

Using \textsc{AutoBench-V}, we conduct an extensive evaluation of nine popular LVLMs in five evaluation capabilities requested (see \autoref{fig:user input}).    
The results show that LVLMs exhibit declining performance as task difficulty rises, with varied performances over distinctive LVLMs. While excelling in high-level understanding, they struggle with fine-grained reasoning, revealing a key area for improvement in future research. Moreover, Some models scored below 25\% on medium and hard tasks, performing worse than random guessing.
To summarize, our key contributions are three-fold:

$\triangleright$ \textbf{An automated LVLM evaluation framework.} The proposed \textsc{AutoBench-V} is the first automated framework for benchmarking LVLMs' capability. The framework uses text-to-image models to generate images based on textual descriptions, and employs LVLM examiners for VQA evaluations. This automation significantly minimizes human involvement, enhancing efficiency and objectivity in the evaluation process.

$\triangleright$ \textbf{Extensive experiments to validate the framework's effectiveness.} We conducted comprehensive experiments, including main evaluations on multiple models, ablation studies on each module, option position bias analyses, and human evaluation. These results confirm the framework’s robustness and effectiveness in evaluating LVLMs.

$\triangleright$ \textbf{In-depth analysis of LVLMs' performance across diverse visual tasks.} Through systematic evaluation with varied user inputs, we obtained many insightful findings. For instance, LVLMs exhibit strong proficiency in abstract conceptual understanding while demonstrating relatively lower performance in concrete visual reasoning tasks. These findings provide insights into the current state of LVLMs, highlighting areas with potential for future development and exploration.

\section{Related Works}
\textbf{Benchmark for LVLMs.}  The emergence of LVLMs greatly promoted the development of multimodal models, showcasing exceptional progress in their multimodal perception and reasoning capabilities. This makes previous benchmarks, which focused on isolated task performance \citep{karpathy2015deep,antol2015vqa} insufficient to provide a comprehensive evaluation. Subsequent studies have introduced various benchmarks to assess LVLMs across a range of multimodal tasks \citep{goyal2017makingvvqamatter,lin2014microsoft,russakovsky2015imagenetlargescalevisual}. However, these benchmarks often lack fine-grained assessments of capabilities and robust evaluation metrics. Hence, recent works \citep{liu2023mmbenchmultimodalmodelallaround,ying2024mmt,fu2024mmecomprehensiveevaluationbenchmark,yu2023mmvetevaluatinglargemultimodal,yu2024mmvetv2challengingbenchmark,zhou2024memo} underscore the critical need for developing advanced, comprehensive benchmarks to more accurately assess multimodal understanding and reasoning capabilities of LVLMs. However, these benchmarks still have various limitations. For example, LVLM-eHub \citep{xu2023lvlmehubcomprehensiveevaluationbenchmark} and LAMM \citep{yin2024lamm} have utilized several classical datasets that are widely recognized but not sufficiently novel for current advancements, overlooking the possibility of data leakage during LVLM training. Hence, MMStar \citep{chen2024we} aims to solve the issure of unnecessary visual content and unintentional data leakage in LVLM training by constructing an elite vision-indispensable dataset. 

\textbf{Automatic benchmarks.}
The significant early advancements in LLMs have driven the development of various benchmarks designed to automate evaluation processes. For example, LMExamQA \citep{bai2024benchmarking} employs the concept of a Language-Model-as-an-Examiner to create a comprehensive and scalable evaluation framework. In addition, \textsc{DyVal} \citep{zhu2023dyval} and \textsc{DyVal2} \citep{zhu2024dynamic} both highlight the importance of dynamic assessment, with \textsc{DyVal} focusing on reasoning tasks and \textsc{DyVal2} adopting a broader psychometric approach. AutoBencher \citep{li2024autobenchercreatingsalientnovel} automates the generation of novel, challenging, and salient datasets for evaluating LLMs, further expanding the scope of automated benchmarking. Other efforts, such as \textsc{UniGen} \citep{wu2024unigenunifiedframeworktextual} and Task Me Anything \citep{zhang2024task}, focus on developing more customized and relevant benchmarks to evaluate LVLM and LLM performance across diverse tasks.

\begin{figure}[t]
    \centering
    \includegraphics[width=\textwidth]{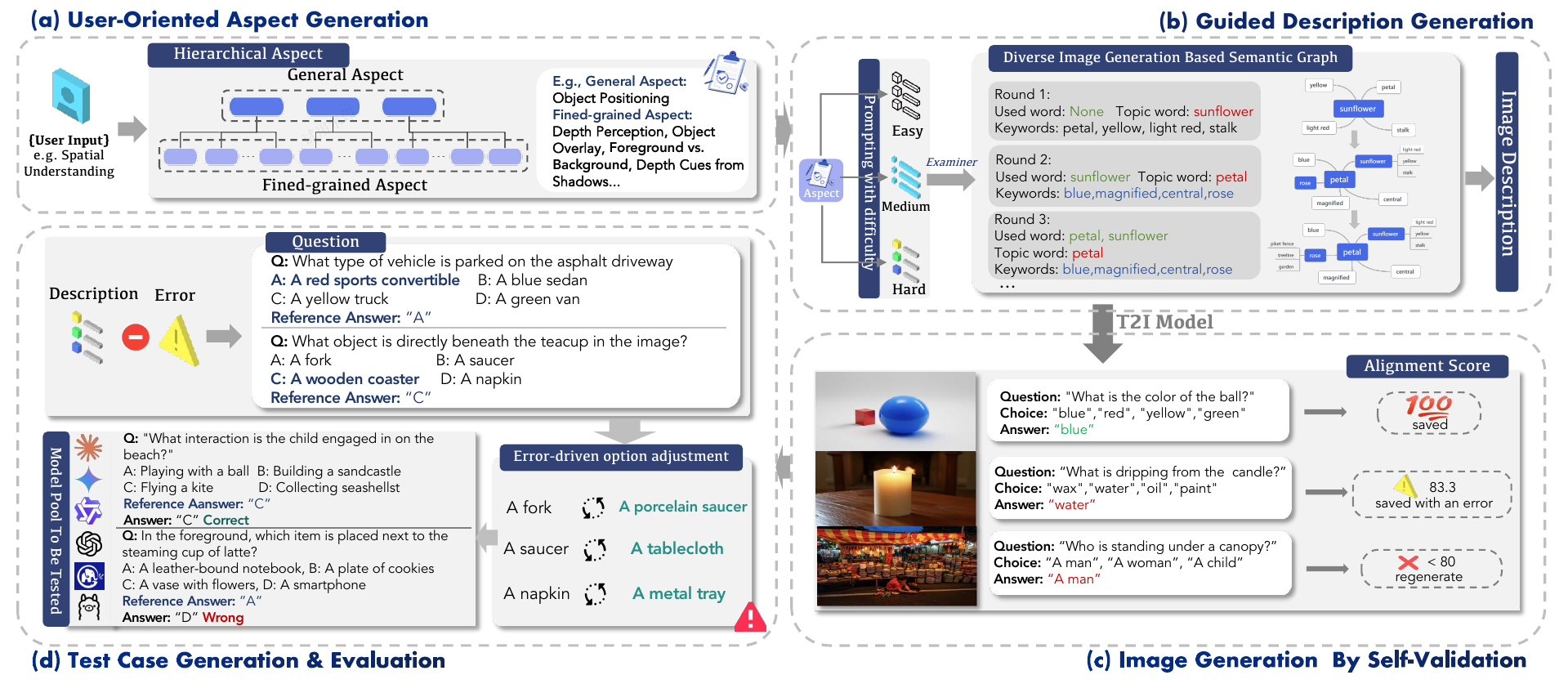}
    \captionsetup{justification=centerlast}
    \caption{A comprehensive overview of the \textsc{AutoBench-V} framework.}
    \label{fig:pipeline}
\end{figure}

\section{The Proposed \textsc{AutoBench-V} Framework}

In this section, we introduce \textsc{AutoBench-V}, a framework for automated benchmarking of LVLMs. 
\textsc{AutoBench-V} makes use of two key components—an LVLM examiner $\mathcal{M}_v$ and a text-to-image model $\mathcal{M}_d$—to support robust and scalable evaluations. 
As illustrated in \autoref{fig:pipeline}, the framework consists of four modules: user-oriented aspect generation, guided description generation, image generation with self-validation, and test-case generation and evaluation. 
\textsc{AutoBench-V} aims to address three primary challenges:


\begin{itemize}[nolistsep, leftmargin=*] 
\item \textbf{Diverse Evaluation Aspects.} To ensure comprehensive benchmarking,
    \textsc{AutoBench-V} covers various capability dimensions, each graded by difficulty. The breadth is achieved via a hierarchical aspect-guided strategy enriched by semantic graph-based prompt generation.
 \item \textbf{Truthful Evaluation Cases.} 
    The framework includes a self-validation step that detects potential errors in generated items. 
    An additional error-control mechanism reinforces factual correctness, improving its reliability.
 \item \textbf{Minimized Evaluation Bias.}
    To reduce bias, \textsc{AutoBench-V} maintains a set of LVLM examiners, limiting self-enhancement bias \citep{ye2024llmjudgebias} that can occur if a single model handles both question generation and responses. 
    Furthermore, error-driven option adjustment thwarts text-only strategies for guessing answers, thereby increasing the difficulty of evaluation items.

\end{itemize}

\subsection{User-Oriented Aspect Generation}

In this part we detail the implementation of \textsc{AutoBench-V} and for the math notations used below are additionally referred to in the Appendix~\ref{app:notation}.

\textbf{User input.} The user input can specify an evaluation target focused on certain aspects of LVLMs' capability. \textsc{AutoBench-V} covers the following key evaluation aspects, which are the most crucial for assessing the capabilities of LVLMs:  
\textbf{\textit{Basic Understanding}}, \textbf{\textit{Spatial Understanding}} \citep{li2023seedbenchbenchmarkingmultimodalllms}, \textbf{\textit{Semantic Understanding}} \citep{meng2024mmiumultimodalmultiimageunderstanding}, \textbf{\textit{Reasoning Capacity}} \citep{liu2023mmbenchmultimodalmodelallaround}, and \textbf{\textit{Atmospheric Understanding}} \citep{geetha2024multimodal}. Notably, the user input is not limited to the above kinds, and   can be customized  as needed.

\textbf{Hierarchical aspect generation.} For each user input, we derive a set of aspects representing specific capability items. For example, as shown in \autoref{fig:user input}, \textit{contextual comprehension} is an aspect under \textit{Basic Understanding}. However, directly generating aspects from user input can lead to excessive repetition, reducing both diversity and reliability by overlapping in semantics and repeatedly evaluating the same capability. To mitigate this, we propose hierarchical aspect generation inspired by the previous study \citep{qin2023toolllm} to constrain the aspect generation process. Formally, given the user input \( q \), we first generate \( n \) general aspects {\small \(\{A^{(g)}_1, A^{(g)}_2, \dots, A^{(g)}_n\}\)} by \(\mathcal{M}_v\), which can be formulated as:  
{\small \(\{A^{(g)}_1, A^{(g)}_2, \dots, A^{(g)}_n\} = \mathcal{M}_v(q)\)}. These general aspects represent high-level evaluation dimensions based on \( q \). Next, for each general aspect \( A^{(g)}_i \), we further generate \( m \) fine-grained aspects {\small \(\{A^{(f)}_{i1}, A^{(f)}_{i2}, \dots, A^{(f)}_{im}\}\)}, where each fine-grained aspect provides more specific criteria related to the general aspect. The fine-grained aspects are also generated by \(\mathcal{M}_v\) and depend on both the user input \( q \) and the corresponding general aspect \( A^{(g)}_i \).  
The fine-grained aspect of generation can be represented as {\small \(\{A^{(f)}_{i1}, A^{(f)}_{i2}, \dots, A^{(f)}_{im}\} = \mathcal{M}_v(q, A^{(g)}_i)\)}. Thus, the hierarchical aspect generation yields a structured set of evaluation aspects (\emph{i.e.}, fine-frained aspect) 
{\small \(\mathcal{A} = \bigcup_{i=1}^{n} \left( \{A^{(g)}_i\} \cup \bigcup_{j=1}^{m} \{A^{(f)}_{ij}\} \right)\)}, where $|\mathcal{A}|$ = $mn$.

\subsection{Diverse Description Generation}
\label{subsection:3.2}

\textbf{Image description with difficulty grading.}
To enable a more comprehensive evaluation, we introduce a difficulty-grading mechanism for the image descriptions, which includes the evaluation cases from different difficulties. We define three difficulty levels: easy, medium, and hard. We show the examples across different difficulties in \autoref{fig:figures}. The difficulty level \( d \) is determined by key factors such as background complexity, element relationships, and the intricacy of textures, which are carefully discussed and designed from the perspective of visual perception. The generation of $\omega$ image descriptions { \(\{\mathcal{T}_{ij1}^d, \mathcal{T}_{ij2}^d, \dots, \mathcal{T}_{ij\omega}^d\}\)} for \(A^{(f)}_{ij}\) at a specific difficulty level \( d \) can be defined as: \(\mathcal\bigcup_{k=1}^{\omega} \{\mathcal{T}_{ijk}^d\} = \mathcal{M}_v(q, A^{(f)}_{ij}, d)\), where \( d \in \{ \text{easy}, \text{medium}, \text{hard} \} \). More details about the difficulty grading are provided in \autoref{Appendix C}.

\textbf{Semantic graph-constraint description generation.}
A key challenge when generating image descriptions at the same difficulty level is minimizing repetitive elements and backgrounds, which can reduce the diversity and generalization of the evaluation. For example, given a user input \( q \) related to spatial understanding, the model \( \mathcal{M}_v \) might tend to produce descriptions centered around urban landscapes, potentially compromising the variety of test cases. To address this, we introduce a description optimization strategy using a semantic graph \citep{quillian1966semantic} to enhance the diversity of image prompts generated by \( \mathcal{M}_v \), with significant results referred to \autoref{fig:topic_words} in \autoref{sec:Appendix A}. Moreover, we show a visualization of specific words in \autoref{fig:semantic_graph} and \autoref{fig:without semantic_graph} in \autoref{sec:Appendix A}. The semantic graph is designed to assist \textsc{AutoBench-V} in avoiding the generation of duplicate descriptions by serving as an iteratively updated structure tailored to prompts within the specified aspect. To illustrate the construction process, without loss of generality. Considering $e'$th iteration of prompt generation, a LLM identified topic word \( t_e \) and a set of $|c|$ related keywords \( K_e = \{ k_{e1}, k_{e2}, \dots, k_{ec} \} \) are selected. These keywords are added as nodes to the semantic graph \( G \), where nodes are connected by edges representing semantic relationships between them. 

Formally, let \( G_e=(V_e, E_e) \) be the semantic graph generated at iteration \( e \), and let \( S_e=\left( V_{e-1} \cup \{ t_e \} \cup K_e \right) \). Then $V_e=S_e \setminus f(S_e)$ represents the node set of topic words and keywords, and \( E_e \) is the set of edges capturing the relationships between them. After each round of prompt generation, we apply a degree-based exclusion algorithm, where the number of excluded nodes is determined by a function \( f(S_e) \). This function defines the number of top-degree nodes to be excluded, allowing flexibility in adjusting how many frequently used words are removed as the iterations progress. Here, the function \( f(S_e) \) can be defined by users' requirements, and in our experiment we define it as $f(S_e) =\underset{V' \subseteq S_e, |V'| = e}{\text{arg max}} \sum_{v \in V'} \deg(v)$, where \( \deg(v) \) represents the degree of node \( v \in V_i \), or it could take a more complex form based on specific conditions.  We mitigate redundancy and promote diversity in the generated prompts by excluding these high-degree nodes, which correspond to the most commonly used words. The function \( f(S_e) \) offers the flexibility to control how aggressively the exclusion process operates based on the round number \( e \).


Overall, the generation of an image description \( \mathcal{T}_{ij}^{e} \) can be formalized as follows:

\begin{equation*}
\bigcup_{k=1}^{\omega} \{\mathcal{T}_{ijk}^{de}\} = {M}_v(q, A^{(f)}_{ij}, V_e, d),
\label{eq:sema_graph}
\end{equation*}
where \( V_e \) represents the refined and diverse set of topic words and keywords after the exclusion mechanism has been applied. We show the detailed procedure in 
Algorithm \ref{Algorithm:DiverseDescriptionGeneration}.

\begin{algorithm}
\caption{\small Diverse Description Generation Strategy}
\label{Algorithm:DiverseDescriptionGeneration}
\renewcommand{\algorithmicrequire}{\textbf{Input:}}
\renewcommand{\algorithmicensure}{\textbf{Output:}}
\small
\begin{algorithmic}[1]
\Require User input \( q \), model \( \mathcal{M}_v \), initial set of topic words and keywords \( V_0 \), exclusion function \( f(S_e) \), number of iterations \( N \)
\Ensure Set of diverse image descriptions \( \{ \mathcal{T}(1), \mathcal{T}(2), \dots, \mathcal{T}(N) \} \)

\State Initialize iteration counter \( e \gets 1 \)
\State Initialize the set of topic words and keywords \( V_1 \gets V_0 \)

\While{$e \leq N$}
    \State Select a topic word \( t_e \) and a set of related keywords \( K_e = \{ k_{e1}, k_{e2}, \dots, k_{ec} \} \)
    \State Form the node set \( S_e = V_{e-1} \cup \{ t_e \} \cup K_e \)
    \State Formulate \( E_e \), where edges represent semantic relationships.
    \State Identify exclusion set 
    \[
    f(S_e) = \underset{V' \subseteq S_e, |V'| = e}{\arg\max} \sum_{v \in V'} \operatorname{deg}(v)
    \]
    \State Update the node set as \( V_e \gets S_e \setminus f(S_e) \)
    \State Set hyperparameters \( d \) and \( \omega \)
    \State Set 
    \[
    \mathcal{T}(e)=\bigcup_{k=1}^{\omega} \{\mathcal{T}_{ijk}^{de}\} = \mathcal{M}_v(q, A^{(f)}_{ij}, \mathcal{D}_{ij}, V_e, d)
    \]
    \State Increment the iteration counter: \( e \gets e + 1 \)
\EndWhile

\Return Set of diverse image descriptions \( \{ \mathcal{T}(1), \mathcal{T}(2), \dots, \mathcal{T}(N) \} \)
\end{algorithmic}
\end{algorithm}

\subsection{Image Generation By Self-Validation}
\label{subsection:3.3}

\textbf{Self-validation.} The image descriptions \(\mathcal{T}_{ij}^d\) and their corresponding aspects \(A^{(f)}_{ij}\) are subsequently provided to the text-to-image model for image generation. At this stage, a potential issue is the possibility of generated images \(\mathcal{I}_{ij}^{d}\) not aligning with the descriptions, due to hallucinations inherent in the text-to-image model \citep{lee2023aligning}. To tackle this issue, drawing inspiration from TIFA \citep{hu2023tifa}, we employed a self-validation process to evaluate the consistency of images with their descriptions via VQA.

In the self-validation process \( \mathcal{F} \), for each image \(\mathcal{I}_{ij}^d\), based on its image description, \( \mathcal{M}_v \) is asked to generate a set of simple questions \( \Phi_{ij}^d = \{ \phi_{ij1}^d, \phi_{ij2}^d, \dots, \phi_{ijp}^d \} \) (\emph{e.g.},\textit{``Is there a wooden chair in the image?''}), where \( p \) denotes the question number to evaluate alignment. The function \( \mathcal{F} \) takes the image \(\mathcal{I}_{ij}^d\), its description \(\mathcal{T}_{ij}^d\), and the set of questions \( \Phi_{ij}^d \) as inputs and outputs an alignment score \( S_{ij}^d \), which is calculated as the ratio of correctly answered questions to the total number of questions, denoted as $S_{ij}^d = \mathcal{F}(\mathcal{I}_{ij}^d, \mathcal{T}_{ij}^d, \Phi_{ij}^d)$.
We set a threshold $\zeta$, where: (\rmnum{1}) If \( S_{ij}^d < \zeta \), the image \(\mathcal{I}_{ij}\) will be reworked in line with the description until it meets the required standard; (\rmnum{2}) If \( \zeta \leq S_{ij}^d < 1 \), the image meets the basic criteria but contains an error \( \mathcal{E}_{ij}^d \), which will be documented for further processing (described in \autoref{subsection:3.4}); and (\rmnum{3}) If \( S_{ij}^d = 1 \), the image is considered to fully align with the description and is deemed acceptable.


\begin{table*}[ht]
\tiny
\centering
\caption{Effectiveness of hierarchical aspect generation under various hyperparameter settings.}
\label{tab:effectiveness}
\renewcommand\arraystretch{1.2} 
\setlength{\tabcolsep}{5pt} 
\scalebox{1.0}{
\begin{tabular}{lccccccccccc}
\toprule[1pt]
\multicolumn{2}{c}{\textbf{m=3, n=5}}  & \multicolumn{2}{c}{\textbf{m=3, n=6}} & \multicolumn{2}{c}{\textbf{m=3, n=7}} & \multicolumn{2}{c}{\textbf{m=4, n=5}}  & \multicolumn{2}{c}{\textbf{m=4, n=6}} & \multicolumn{2}{c}{\textbf{m=4, n=7}} \\
\cmidrule(lr){1-2} \cmidrule(lr){3-4} \cmidrule(lr){5-6} \cmidrule(lr){7-8} \cmidrule(lr){9-10} \cmidrule(lr){11-12}
\texttt{Raw} & \texttt{+Hierarchy}  & \texttt{Raw} & \texttt{+Hierarchy}  & \texttt{Raw}  & \texttt{+Hierarchy}  & \texttt{Raw} & \texttt{+Hierarchy} & \texttt{Raw} & \texttt{+Hierarchy}  & \texttt{Raw}  & \texttt{+Hierarchy} \\
\midrule
\midrule
0.767  & 0.778 (\textcolor{green!60!black}{1.4\% $\uparrow$})   
& 0.773 & 0.780 (\textcolor{green!60!black}{1.0\% $\uparrow$})  
& 0.779 & 0.825 (\textcolor{green!60!black}{5.9\% $\uparrow$})  
& 0.780 & 0.790 (\textcolor{green!60!black}{1.3\% $\uparrow$})   
& \textbf{0.786} & \textbf{0.849} (\textcolor{green!60!black}{\textbf{10.2\%} $\uparrow$})  
& 0.798 & 0.842 (\textcolor{green!60!black}{5.5\% $\uparrow$})  \\
\bottomrule[1pt]
\end{tabular}
}
\end{table*}

\begin{figure*}[t]
    \centering
    \includegraphics[width=1.0\textwidth]{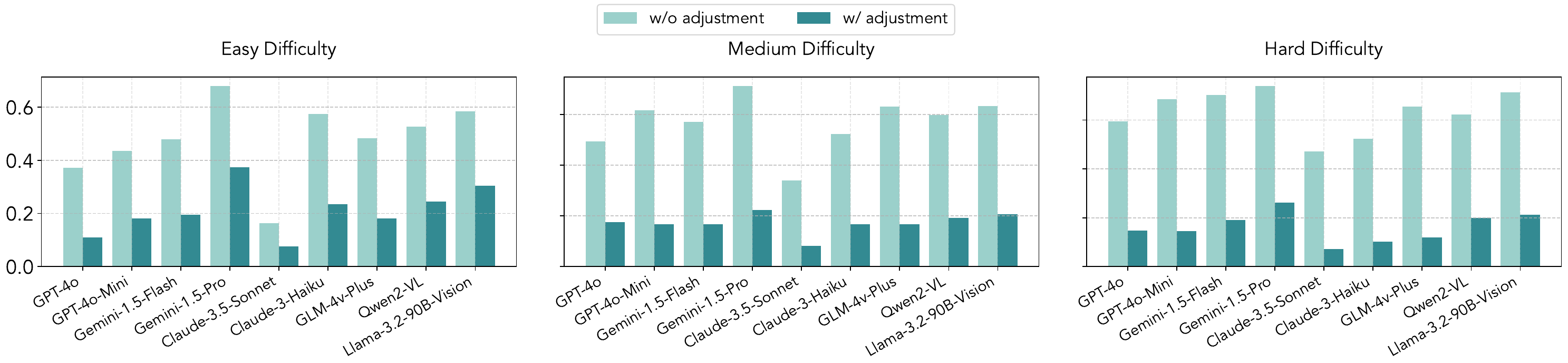}
    \caption{Models' accuracy (here we take ``basic understanding'' as an example) before and after the option adjustment, without image input (only guess the answer by textual information).}
    \label{fig:answer_leakage}
\end{figure*}


\subsection{Test Case Generation \& Evaluation}
\label{subsection:3.4}

\textbf{Q\&A generation with error control.} To enhance question generation accuracy, especially for addressing potential image flaws, we propose an error control mechanism. While self-validation helps, not all images can be guaranteed flawless. To mitigate biases from relying on a single examiner LVLM \citep{zhang2024debiasingmultimodallargelanguage}, we incorporate a diverse set of examiners, inspired by prior work \cite{bai2024benchmarking}. This approach reduces self-enhancement bias \cite{ye2024llmjudgebias} and promotes greater diversity in evaluation.

When generating questions, we will only include the image description \(\mathcal{T}_{ij}^d\) and any identified defects \(\mathcal{E}_{ij}^d\) in the input to the examiners \(\mathcal{M}_v^k\), where \(k\) represents the index of the selected examiner from the set of available examiners. Each time a question is generated, an examiner \(\mathcal{M}_v^k\) will be randomly selected. The function \( \mathcal{M}_v^k \) generates the question \( Q_{ij}^d \) based on the image description and errors, denoted as $Q_{ij}^d = \mathcal{M}_v^k(\mathcal{T}_{ij}^d, \mathcal{E}_{ij}^d)$.

Moreover, the options for each question will also be generated by a randomly selected examiner from the set. For each image, we will provide a related question \( Q_{ij}^d \) (\emph{e.g.}, \textit{multiple-choice or true/false}). These questions, along with the accompanying images, will be presented to the LVLMs under evaluation for their response.

\textbf{Error-driven option adjustment.} A key challenge in evaluating LVLMs is the generation of distractor options that are both plausible and challenging. Distractors generated by LVLMs are often trivial, leading to inflated performance, even under high-difficulty settings or purely text-based evaluations, as models can infer correct answers without relying on visual inputs \citep{chen2024we}. Attempts to directly create harder options often result in low-quality, overly obscure distractors.

To address this, we propose \textit{error-driven option adjustment}, a mechanism to improve distractor difficulty while preserving quality. For a given question \( Q_{ij}^d \), with correct option \( O_{ij}^c \) and distractors \( O_{ij}^f = \{ O_{ij,1}^f, \dots, O_{ij,k}^f \} \). First, the randomly selected LVLM examiner is given \( Q_{ij}^d \) and options \( \{ O_{ij}^c \} \cup O_{ij}^f \), but the correct option \( O_{ij}^c \) is intentionally mislabeled as incorrect. The model generates a plausible “correct answer” \( O_{ij}^{f*} \), given by $O_{ij}^{f*} = \mathcal{M}_v^k(Q_{ij}^d, O_{ij}^c)$. This generated \( O_{ij}^{f*} \) replaces one or more original distractors in \( O_{ij}^f \), resulting in an updated set \( O_{ij}^{f'} \). This adjustment increases the difficulty of distinguishing the correct answer from distractors, as \( O_{ij}^{f'} \) now contains plausible alternatives generated by the model itself under adversarial conditions.
\textbf{Evaluation.} The response \(\mathcal{P}_{ij}^d\) from the tested LVLMs was compared to the reference answer \(A_{ij}^d\) to determine accuracy. The overall accuracy was calculated as the percentage of correctly answered questions. By given the reference answer, we use LLM-as-a-Judge \cite{zheng2023judging} to label the responses.

\section{Experiments}

\subsection{Experiment Setup}

\begin{table*}[t]
\centering
\renewcommand\arraystretch{1.0}
\small
\setlength{\tabcolsep}{3pt} 
\caption{Performance (accuracy) of all models on five user inputs and three difficulty levels (red and blue represent the highest and lowest scores, respectively, in each row). Claude-3.5 is Claude-3.5-Sonnet and Llama-3.2-90B is Llama-3.2-90B-Vision.}
\label{tab:overall}
\scalebox{0.9}{
    \begin{tabular}{l l>{\centering\arraybackslash}p{1.4cm} >{\centering\arraybackslash}p{1.4cm} >{\centering\arraybackslash}p{1.4cm} >{\centering\arraybackslash}p{1.4cm} >{\centering\arraybackslash}p{1.4cm} >{\centering\arraybackslash}p{1.4cm} >{\centering\arraybackslash}p{1.4cm} >{\centering\arraybackslash}p{1.4cm} >{\centering\arraybackslash}p{1.4cm}}
    \toprule[1pt]
    \multirow{2}{*}{\textbf{User Input}} & \multirow{2}{*}{\textbf{Difficulty}} & \multicolumn{9}{c}{\textbf{Model}} \\ 
    \cmidrule(lr){3-11}
    & & \textbf{GPT-4o} & \textbf{GPT-4o mini} & \textbf{Gemini-1.5-Flash} & \textbf{Gemini-1.5-Pro} & \textbf{Claude-3.5} & \textbf{Claude-3-Haiku} & \textbf{GLM-4v-Plus} & \textbf{Qwen2-VL} & \textbf{Llama-3.2-90B} \\
    \midrule
    \multirow{3}{*}{\textsc{Basic.}} 
      & Easy   & 77.03\% & 72.73\% & 78.47\% & 78.47\% & \cellcolor[HTML]{ffccd5}84.69\% & 70.81\% & 74.64\% & 76.56\% & \cellcolor[HTML]{ade8f4}50.72\% \\
      & Medium & 56.65\% & 46.35\% & 44.21\% & 51.93\% & \cellcolor[HTML]{ffccd5}57.51\% & 44.64\% & 47.64\% & 53.88\% & \cellcolor[HTML]{ade8f4}32.19\% \\
      & Hard   & 41.15\% & 30.53\% & 34.07\% & 39.82\% & \cellcolor[HTML]{ffccd5}45.58\% & 31.42\% & 34.07\% & \cellcolor[HTML]{ffccd5}45.58\% & \cellcolor[HTML]{ade8f4}29.78\% \\
    \midrule
    \multirow{3}{*}{\textsc{Spatial.}} 
      & Easy   & 59.22\% & 44.69\% & 60.00\% &  60.56\%  & \cellcolor[HTML]{ffccd5}65.00\% & 42.78\% & 57.22\% & 62.78\% &  \cellcolor[HTML]{ade8f4}36.31\%  \\
      & Medium & 34.50\% & \cellcolor[HTML]{ade8f4}22.71\% & 29.69\% &  \cellcolor[HTML]{ffccd5}41.48\%   & 38.86\% & 25.33\% & 31.44\% & 36.68\% &   24.45\%    \\
      & Hard   & 21.46\% & 15.14\% & 23.32\% & \cellcolor[HTML]{ffccd5}27.35\%  & 26.01\% & 15.84\% & 17.49\% & 25.91\% &  \cellcolor[HTML]{ade8f4}14.55\%     \\
    \midrule
    \multirow{3}{*}{\textsc{Seman.}} 
      & Easy   & \cellcolor[HTML]{ffccd5}73.10\% & 64.62\% & 68.53\% &  70.56\%  & 72.59\% & 55.84\% & 62.94\% & 66.33\% &  \cellcolor[HTML]{ade8f4}42.64\%     \\
      & Medium & \cellcolor[HTML]{ffccd5}61.90\% & 46.75\% & 54.11\% &  58.44\%  & \cellcolor[HTML]{ffccd5}61.90\% & 46.75\% & 51.95\% & 61.04\% &  \cellcolor[HTML]{ade8f4}39.57\% \\
      & Hard   & 43.64\% & 33.79\% & 44.55\% & \cellcolor[HTML]{ffccd5}47.27\% & 45.45\% & \cellcolor[HTML]{ade8f4}26.36\% & 36.36\% & 44.55\% &  30.14\%  \\
    \midrule
    \multirow{3}{*}{\textsc{Reason.}} 
      & Easy   & 57.06\% & 46.20\% & 56.40\% &  \cellcolor[HTML]{ffccd5}57.56\% & 57.00\% & 36.63\% & 46.51\% & 53.49\% &  \cellcolor[HTML]{ade8f4}36.26\% \\
      & Medium & 47.16\% & 35.22\% & 40.43\% & 47.39\%  & 46.09\% & \cellcolor[HTML]{ade8f4}25.22\% & 36.96\% & \cellcolor[HTML]{ffccd5}48.03\% &  26.64\%     \\
      & Hard   & 38.21\% & 30.66\% & 21.43\% & 35.71\%  & 36.59\% & \cellcolor[HTML]{ade8f4}14.29\% & 24.53\% & \cellcolor[HTML]{ffccd5}39.15\% &   25.00\%    \\
    \midrule
    \multirow{3}{*}{\textsc{Atmos.}} 
      & Easy   & 60.20\% & 55.94\% & 58.91\% &  \cellcolor[HTML]{ffccd5}66.83\%  & 60.40\% & 48.02\% & 52.48\% & 56.72\% &  \cellcolor[HTML]{ade8f4}36.63\%     \\
      & Medium & 39.73\% & 30.67\% & 39.21\% &  39.65\%  & \cellcolor[HTML]{ffccd5}44.93\% & \cellcolor[HTML]{ade8f4}27.75\% & 33.48\% & 39.21\% & 28.63\%     \\
      & Hard   & 33.79\% & 18.26\% & 28.64\% &  30.45\% & \cellcolor[HTML]{ffccd5}34.55\% & \cellcolor[HTML]{ade8f4}15.00\% & 23.18\% & 31.05\% &   22.73\%  \\
    \midrule
    \multicolumn{2}{l}{\textsc{Average}} 
      & 49.65\% & 39.63\% & 45.46\% & 50.23\% & \cellcolor[HTML]{ffccd5}51.81\% & 35.11\% & 42.06\% & 49.40\% & \cellcolor[HTML]{ade8f4}31.75\% \\
    \bottomrule[1pt]
    \label{tab:combined_scores}
    \end{tabular}
}
\end{table*}

\textbf{Models.} In evaluating LVLMs, we selected nine representative models: {GPT-4o}, {GPT-4o mini} \citep{openai2024gpt4technicalreport}, {Claude-3.5-Sonnet}, {Claude-3-Haiku} \citep{Claude-3-family}, {Gemini-1.5-Flash} \citep{Gemini-1-5-flash}, {Gemini-1.5-Pro} \citep{geminipro}, {GLM-4v-Plus} \citep{glm2024chatglmfamilylargelanguage}, and the open-source models: {Qwen2-VL} \citep{qwen}, {Llama-3.2-90B-Vision} \citep{llama3.2}, detailed in \autoref{tab:model selection}. These advanced models exhibit exceptional image understanding. Some well-known open-source models, such as {Llava-1.6} \citep{liu2023visualinstructiontuning} and {MiniGPT-4} \citep{zhu2023minigpt}, were tested and found to perform poorly and exclude finally from the model list. We include {GPT-4o}, {Gemini-1.5-Pro}, and {Claude-3.5-Sonnet} in the examiner pool as examiner models for generation due to their strong overall performance. The descriptions were then passed to {Flux-1.1-Pro} \citep{flux2023}, a text-to-image model known for outstanding image generation. We also experimented with other text-to-image models \citep{rombach2022high,podellsdxl,betker2023improving} but did not include them because of their suboptimal performance. 

\textbf{Hyperparameter Setting.} We set $\displaystyle n = 4$ for the number of general aspects and $\displaystyle m = 6$ for the number of fine-grained aspects, as this configuration yields the highest diversity in the generated aspects as illustrated in \autoref{tab:effectiveness}, allowing for a broader range of scenes and elements. We set $\omega$  = 10, namely 10 pictures for each fine-grained aspect. Therefore, we evaluate 720 images for each user input (with each user input having three difficulty levels). For easy difficulty, we set the self-validation threshold $\zeta_e = 1$ since the scenes are simpler and contain fewer elements, which justifies a higher threshold.  For the medium and hard difficulty levels, the images contain more elements, so we lower the thresholds to $\zeta_m = \zeta_h = 0.8$ to avoid compromising efficiency. It balances minimizing errors and avoiding excessive image rejection; a smaller threshold risks inaccuracies, while a larger one may discard too many images. The error control mechanism ensures high-quality question generation.

\begin{table}[t]
    \centering
    \small
    \renewcommand\arraystretch{1.0}
    \setlength{\tabcolsep}{12pt} 
    \caption{Average accuracy for various user inputs at different difficulty levels (red and blue represent the highest and lowest scores, respectively, in each column).}
    \scalebox{1.0}{
    \begin{tabular}{lcccc}
    \toprule[1pt]
    \textbf{User Input} & \textbf{Easy} & \textbf{Medium} & \textbf{Hard} &\textbf{Average}\\
    \midrule
    \textbf{\textsc{Basic.}}       & \cellcolor[HTML]{ffccd5}73.79\% & 48.33\% & 36.88\% & \cellcolor[HTML]{ffccd5}53.00\%\\
    \textbf{\textsc{Spatial.}}     & 54.28\% & \cellcolor[HTML]{ade8f4}31.68\% & \cellcolor[HTML]{ade8f4}20.78\% & \cellcolor[HTML]{ade8f4}35.58\%\\
    \textbf{\textsc{Seman.}}    & 64.13\% & \cellcolor[HTML]{ffccd5}53.60\% & \cellcolor[HTML]{ffccd5}39.12\% &52.28\%\\
    \textbf{\textsc{Reason.}}      & \cellcolor[HTML]{ade8f4}49.67\% & 39.24\% & 29.51\% &39.47\%\\
    \textbf{\textsc{Atmos.}}  & 55.12\% & 35.92\% & 26.41\% & 39.15\%\\
    \bottomrule[1pt]
    \label{tab:average_scores_transposed}
    \end{tabular}
    }
\end{table}

\subsection{Main Results}

\textit{\ul{Performance drops significantly with increasing difficulty.}}
As shown in \autoref{tab:combined_scores} and \autoref{tab:average_scores_transposed}, across all models and user inputs, scores decrease as the difficulty progresses from easy to medium to hard. In the basic understanding, the average score across all models drops from approximately 73.79\% at easy to 36.88\% at hard. This trend is consistent across the \autoref{tab:combined_scores}, where the hard is noticeably shorter than the easy and medium.

\textit{\ul{Basic understanding is the simplest task, while spatial understanding represents the most challenging one:}} Most models perform better in \textit{basic understanding} compared to \textit{spatial understanding}. For example, the average score for basic understanding at easy difficulty is 73.79\%, while \textit{spatial understanding} achieves lower averages of 54.28\%. This indicates the imbalance of current LVLMs' capabilities and emphasizes the importance of holistic improvement.

\textit{\ul{Some models achieve scores below 25\% on tasks with medium and hard difficulty, which suggests that the model performs worse than random guessing.}} As shown in \autoref{tab:combined_scores}, GPT-4o Mini, Gemini-1.5-Flash, Claude-3.5-Haiku, GLM-4v-Plus, and Llama-3.2-90B-Vision achieved scores below 25\% on hard tasks involving spatial understanding and reasoning capacity. This indicates that AutoBench-V effectively challenges the evaluated models, mitigating the answer leakage issue where models rely solely on textual information to answer correctly (we will discuss this in detail in \autoref{sec:exploratory}).

\begin{table}[t]
\centering
\renewcommand\arraystretch{1.0}
\small
\setlength{\tabcolsep}{6pt}
\caption{Average performance (Accuracy) of all models at different difficulty levels (red and blue represent the highest and lowest scores, respectively, in each column).}
\scalebox{1.0}{
    \begin{tabular}{l >{\centering\arraybackslash}p{1.5cm} >{\centering\arraybackslash}p{1.5cm} >{\centering\arraybackslash}p{1.5cm}}
    \toprule[1pt]
    \textbf{Model} & \textbf{Easy} & \textbf{Medium} & \textbf{Hard} \\
    \midrule
\textbf{GPT-4o}              & 65.32\% & 47.99\% & 35.65\% \\
\textbf{GPT-4o mini}         & 56.83\% & 36.34\% & 25.68\% \\
\textbf{Gemini-1.5-Flash}    & 64.46\% & 41.53\% & 30.40\% \\
\textbf{Gemini-1.5-Pro}      & 66.79\% & 47.78\% & 36.12\% \\
\textbf{Claude-3.5-Sonnet}   & \cellcolor[HTML]{ffccd5}67.93\% & \cellcolor[HTML]{ffccd5}49.86\% & \cellcolor[HTML]{ffccd5}37.63\% \\
\textbf{Claude-3-Haiku}      & 50.82\% & 33.94\% & 20.58\% \\
\textbf{GLM-4v-Plus}         & 58.76\% & 40.29\% & 27.13\% \\
\textbf{Qwen2-VL}            & 63.17\% & 47.77\% & 37.25\% \\
\textbf{Llama-3.2-90B-Vision}  & \cellcolor[HTML]{ade8f4}40.51\%  & \cellcolor[HTML]{ade8f4}30.30\%  & \cellcolor[HTML]{ade8f4}24.44\%\\
    \bottomrule[1pt]
    \end{tabular}
}
\label{tab:average_scores}
\end{table}

\subsection{Exploratory Experiments}
\label{sec:exploratory}

We aim to develop a deeper understanding of the following questions at this section: 
\begin{itemize}[nolistsep, leftmargin=*]
    \item \textbf{Q1:} What impact does error-driven option adjustment have on visual question answering?
    \item \textbf{Q2:} What are the advantages of multi-evaluator assessment compared to single-evaluator evaluation? 
    \item \textbf{Q3:} Does the position of correct options affect model performance?
    \item \textbf{Q4:} Is the ranking of the same model consistent across different user inputs and different levels of difficulty?
\end{itemize}


\begin{figure}
\centering
    \includegraphics[width=0.7\linewidth]{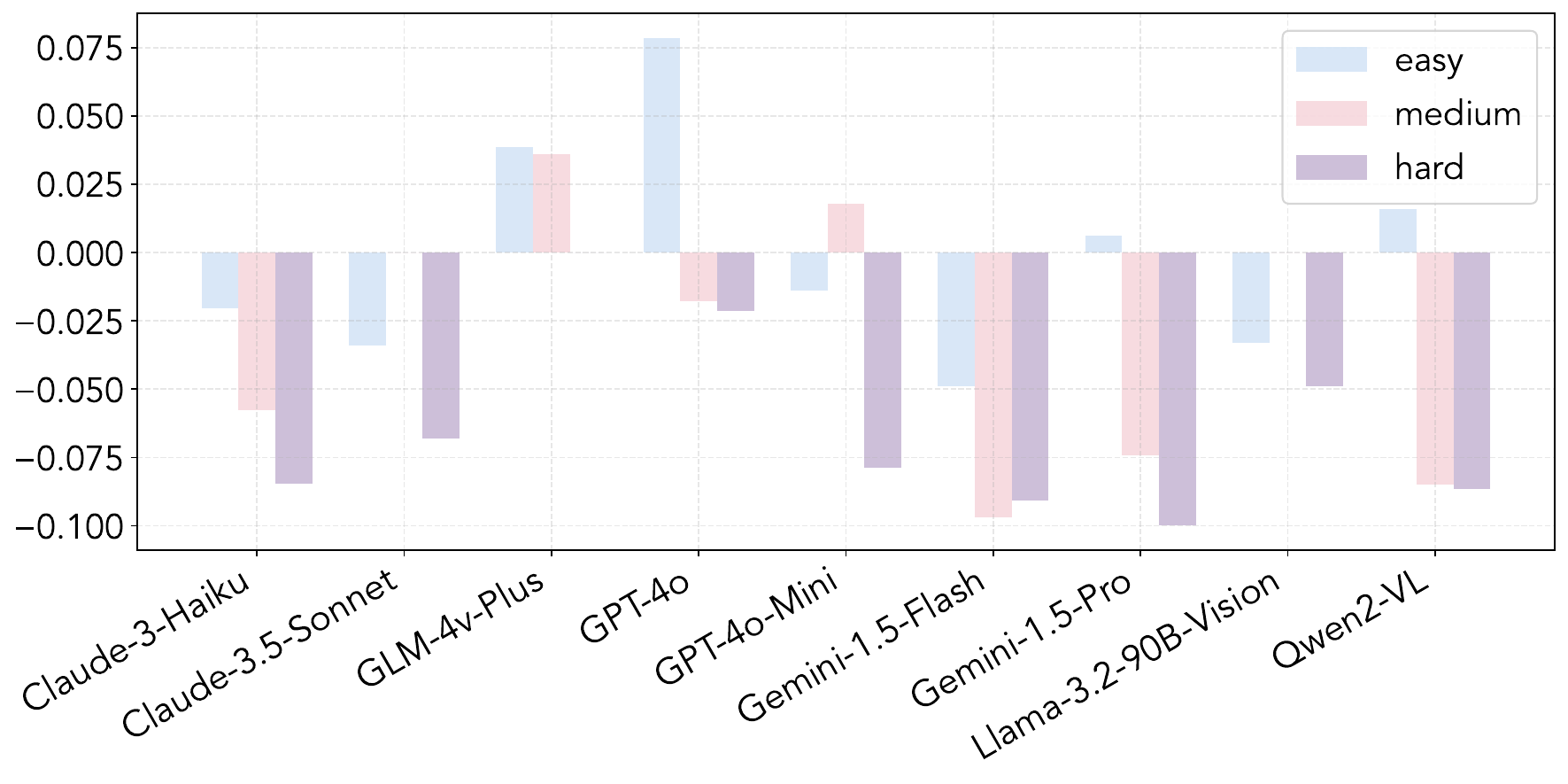}
    \caption{Comparison of answer distribution under position bias conditions. Correct answers at A v.s. correct answers are evenly distributed at A,B,C,D.}
    \label{fig:deviation rate}
\end{figure}

\textit{\ul{Unnecessary visual information was significantly reduced after the option adjustment.}} As illustrated in \autoref{fig:answer_leakage}, in the absence of image input, the scores of various models before option adjustment were concentrated within the range of 40 to 60. Interestingly, without option adjustment, the ``guess accuracy'' significantly increases with the difficulty level. For instance, GPT-4o can guess less than 40\% answers correctly on the easy level while achieving a guessing accuracy of 60\% on the hard level. After option adjustment, the performance of all models declined significantly, demonstrating that error-driven option adjustment effectively mitigates answer leakage in the questions.

\begin{wrapfigure}{r}{8cm}
    \includegraphics[width=7.5cm]{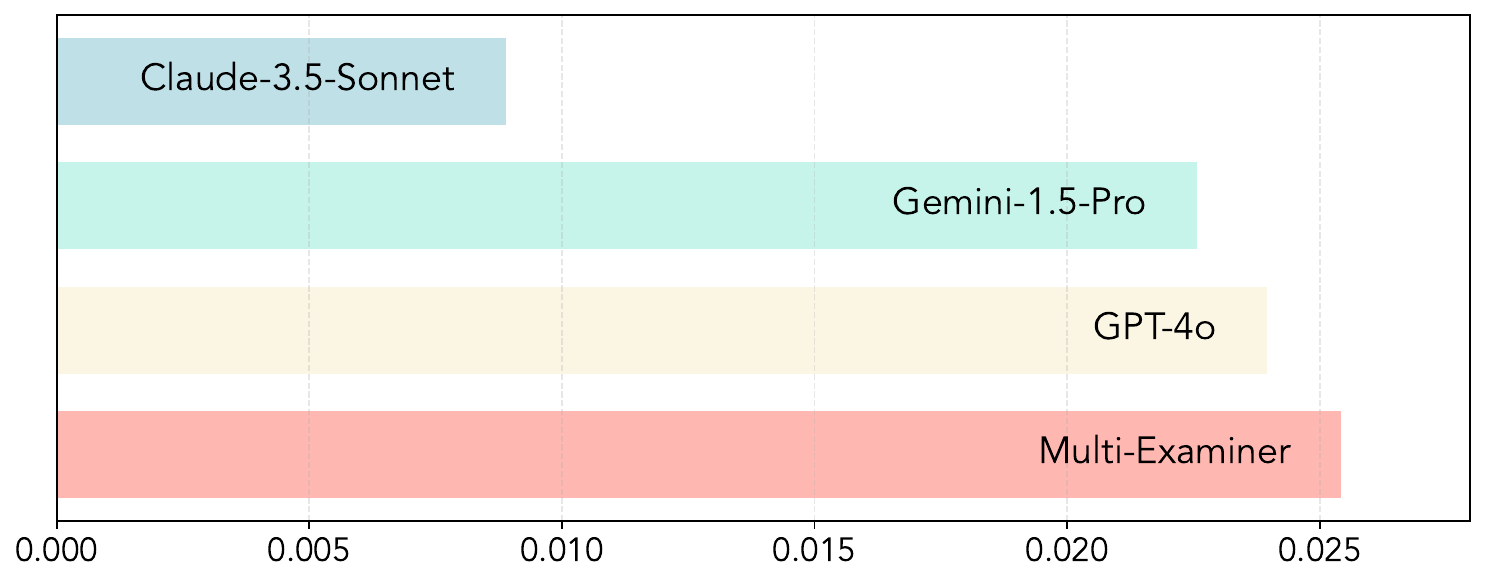}
    \caption{The score variance of {GPT-4o}, {Gemini-1.5-Pro}, and {Claude-3.5-Sonnet} when using different models as examiners.}
    \label{fig:examiner_variance}
\end{wrapfigure}

\textit{\ul{Under the setting of multiple examiners, the performance differences among models become more pronounced.}} As shown in \autoref{fig:examiner_variance}, when multiple evaluators are employed, the variance in the performance of the three models is significantly larger. This suggests that using a single model for evaluation may reduce the difference gap between models, leading to an ``evaluation bottleneck'' while the multi-examiner can mitigate this to achieve a more diverse assessment and lead to a larger performance gap across models.

\begin{figure}
\centering
    \includegraphics[width=0.7\linewidth]{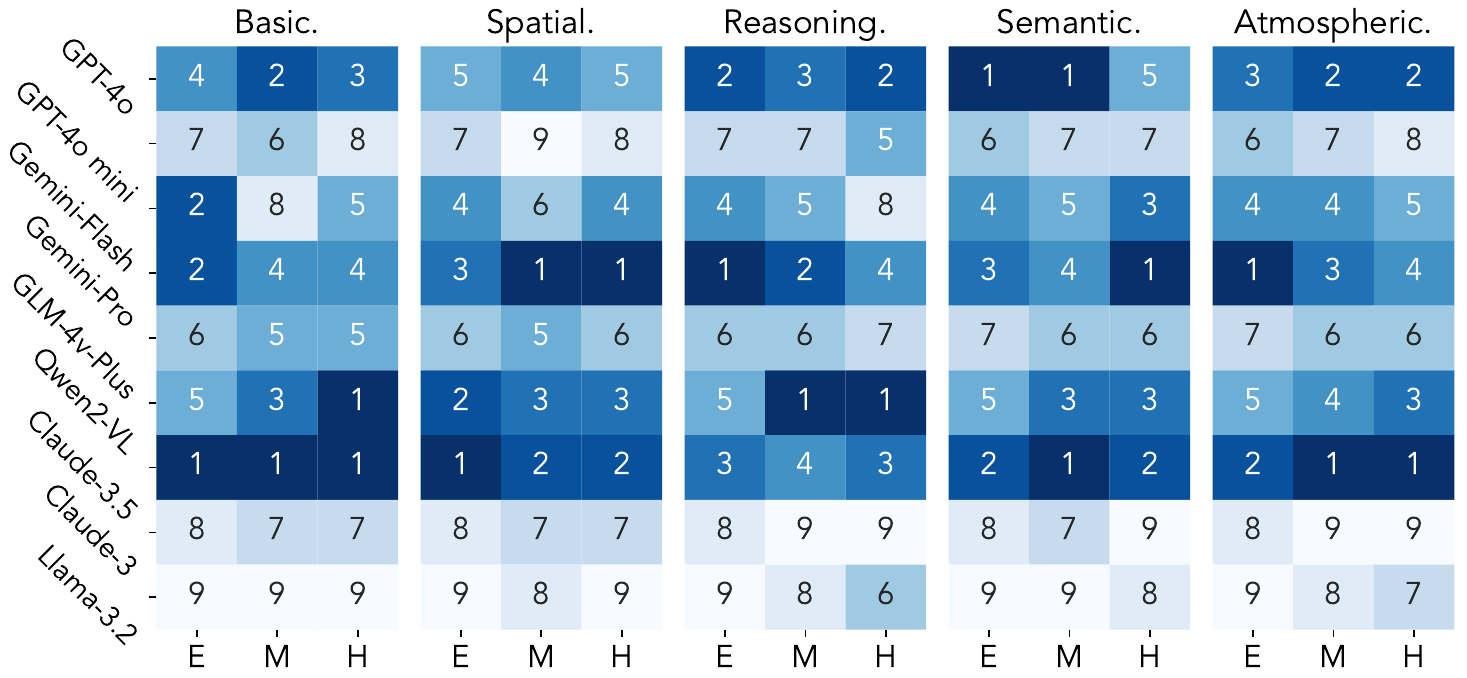}
    \captionsetup{justification=centerlast}
    \caption{The model performance ranking given five user inputs under different difficulty levels.}
    \label{fig:heatmap}
\end{figure}

\textit{\ul{The position of the correct options can influence model performance.}}
To avoid potential position bias in the evaluation process, we ensured an even distribution of correct options. To further explore the possible effects of position bias on model performance, we conducted a case study where all correct answers were intentionally placed in option A and compared it to the evenly distributed case (25\% per option) to assess the necessity of maintaining even distribution, as shown in \autoref{fig:deviation rate}. The deviation rate was calculated using the formula \( R = \frac{S_A - S_U}{S_U} \), where \( S_A \) is the score with correct answers concentrated in option A, and \( S_U \) is the score under evenly distributed conditions.
As illustrated in \autoref{fig:deviation rate}, most models exhibit decreased performance after the answer distribution is balanced, indicating a strong preference for selecting option A.

\textit{\ul{Different models have distinct performance patterns under various difficulties.}} As shown in \autoref{fig:heatmap}, Claude-3.5-Sonnet achieves the best performance, consistently ranking highest across most tasks, while Llama-3.2-90B-Vision performs the weakest, frequently ranking last. Interestingly, models like Qwen2-VL rank higher on hard tasks than on easy ones, whereas models like Gemini-1.5-Pro excel on easy tasks but decline slightly on hard ones. These findings highlight the importance of difficulty-based evaluation and demonstrate \textsc{AutoBench-V}’s effectiveness in revealing such performance imbalances.



\subsection{Human Evaluation}
We conducted human evaluations to assess the alignment between the generated questions and the reference answers. The evaluations were performed using a metric we refer to as the \textit{alignment rate}, defined as the proportion of aligned samples out of the total. Detailed procedures for the human evaluation can be found in \autoref{humaneval detail}.

The results show that the model achieves high alignment across difficulty levels: 95.20\% for Easy tasks, 88.13\% for Medium, and 84.55\% for Hard. This demonstrates the model’s ability to generate well-aligned question-answer pairs, excelling in simpler tasks while maintaining strong performance under higher complexity, highlighting its effectiveness in producing accurate, contextually appropriate outputs for visual tasks.

\section{Conclusion}
In this work, we introduce \textsc{AutoBench-V}, an automated framework designed to benchmark LVLM. It integrates a series of innovative modules that ensure diversity and reliability in dataset generation and evaluation. Extensive experiments demonstrate its robustness and unbiased process, offering a solid foundation for future research.

\newpage

\bibliography{arxiv_paper}

\newpage
\appendix
\onecolumn
\section{Details of Experiment Setting \& Result}
\label{sec:Appendix A}

\textbf{Model Selection.} The details of models selected in our experiments are shown in \autoref{tab:model selection}.


\textbf{Alignment Evaluation.} Inspired by TIFA \citep{hu2023tifa}, we generated consistency tests for images across 12 aspects:\textit{object, human, animal, food, activity, attribute, counting, color, material, spatial, location, shape, other}. 


\begin{figure}[ht]
    \centering
    \includegraphics[width=1.0\textwidth]{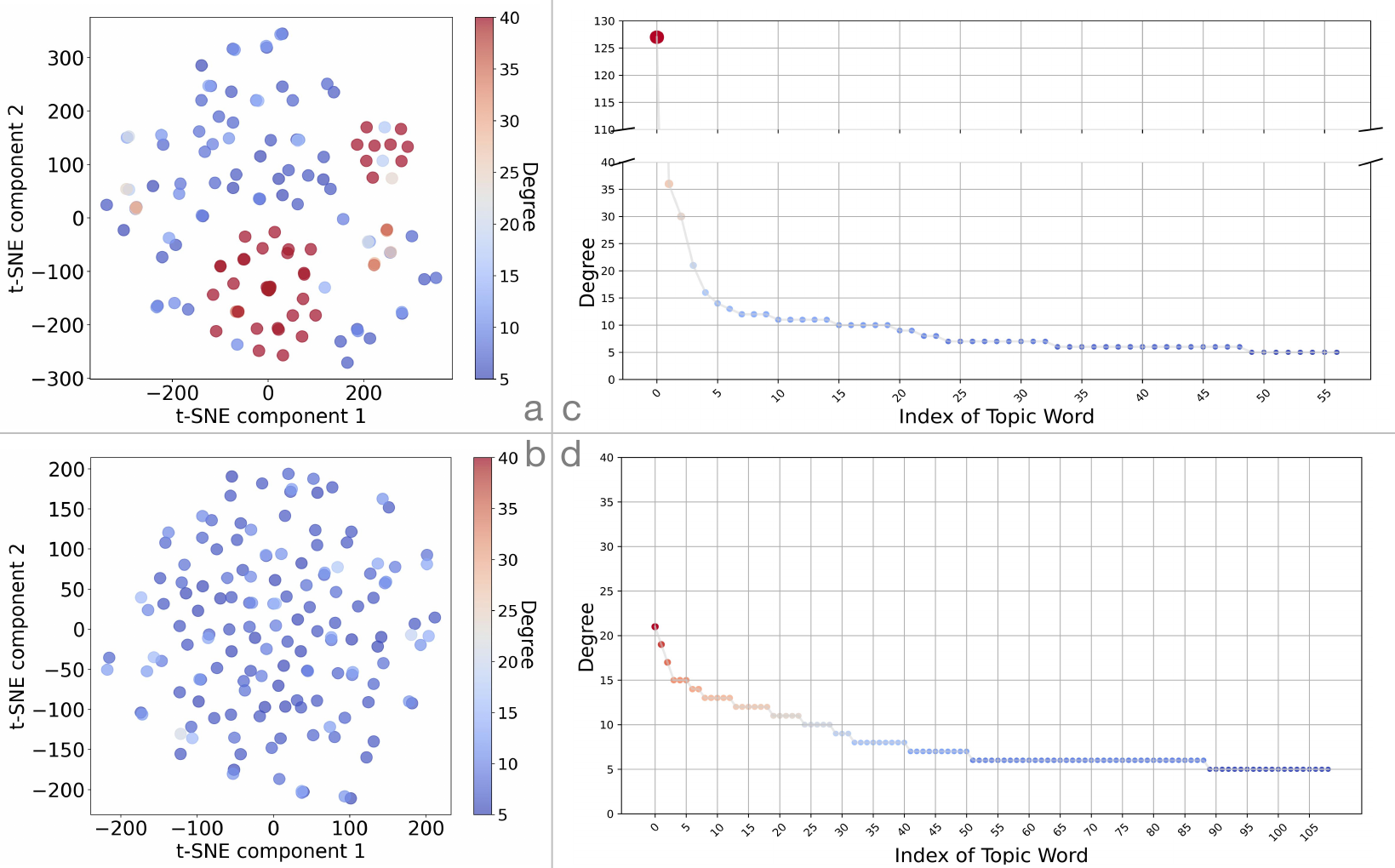}
    \caption{Visualization of image topic words. Topic words are converted into vectors using {bge-large-en-v1.5} \citep{xiao2024cpackpackagedresourcesadvance}, then perform dimensionality reduction via t-SNE \citep{van2008visualizing}. Topic word distribution without semantic graph (a)(c) and with semantic graph (b)(d). It can be seen that with the semantic graph the diversity of topic words increases and the over-reliance on high-degree words is reduced.}
    \label{fig:topic_words}
\end{figure}

\begin{figure}[ht]
    \centering
    \includegraphics[width=0.95\textwidth]{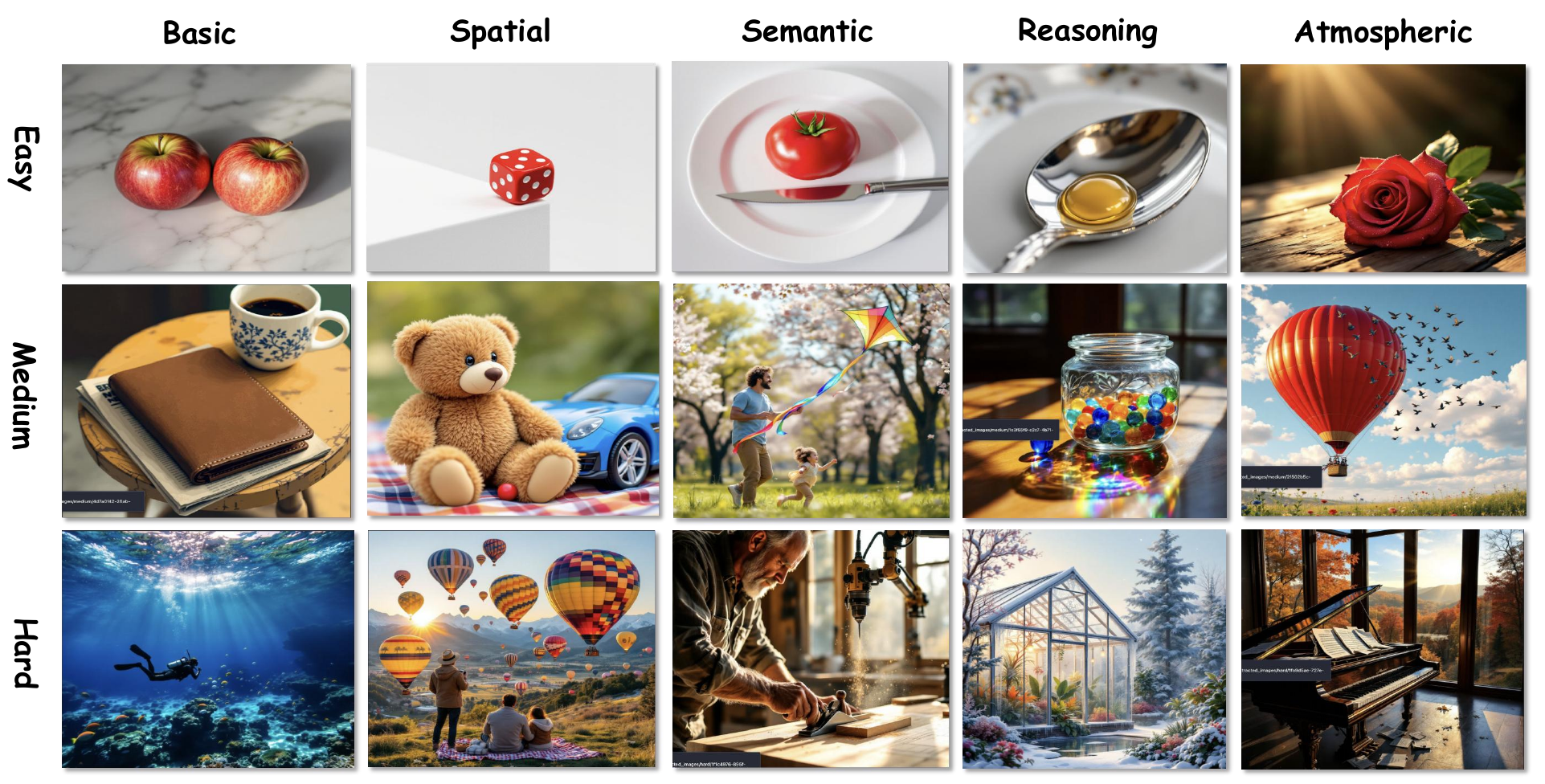}
    \captionsetup{justification=centerlast}
    \caption{Images examples corresponding to different user inputs at varying difficulty levels.}
    \label{fig:figures}
\end{figure}

\begin{figure}[t]
    \centering
    \includegraphics[width=0.7\linewidth]{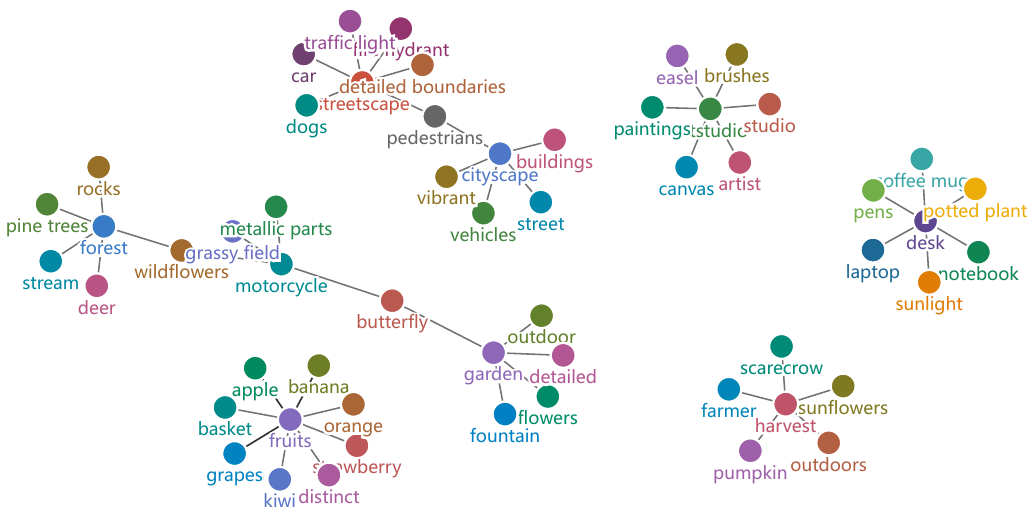}
    \captionsetup{justification=centerlast}
    \caption{Topic words visualization using semantic graph under basic understanding.}
    \label{fig:semantic_graph}
\end{figure}

\begin{figure}[t]
    \centering
    \includegraphics[width=0.50\linewidth]{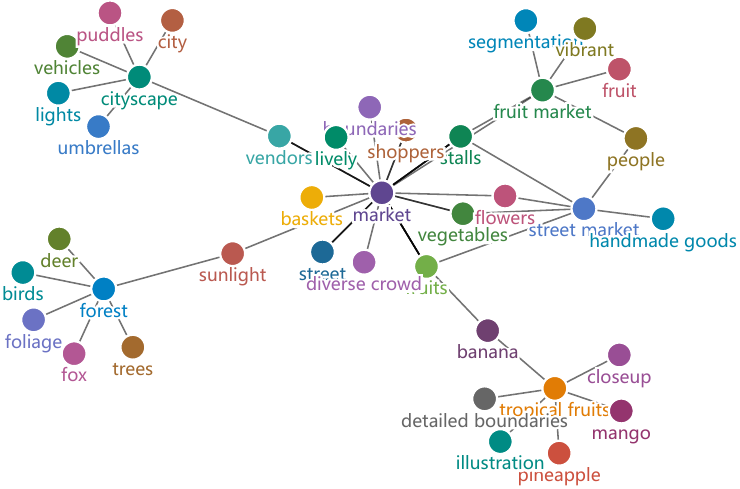}
    \captionsetup{justification=centerlast}
    \caption{Topic words visualization without using semantic graph under basic understanding.}
    \label{fig:without semantic_graph}
\end{figure}

\begin{table*}[ht]
    \centering
    \small
    \captionsetup{justification=centerlast}
    \caption{Model names, Creators, whether it is open source, and their purpose.}
    \setlength{\tabcolsep}{10pt} 
    \renewcommand\arraystretch{1.3}
    \label{tab:model selection}
    \begin{tabular}{cccc}
    \toprule[1pt]
    \multicolumn{1}{c}{\textbf{Model}} & \textbf{Creator} & \multicolumn{1}{c}{\textbf{Open-Source}} & \textbf{Purpose} \\
    \midrule
    {GPT-4o}            & \multirow{2}{*}{OpenAI}          & \textcolor{red}{\faTimesCircleO} & Examiner\&Candidate \\
    {GPT-4o mini}       &                                 & \textcolor{red}{\faTimesCircleO} & Candidate \\
    \hline
    {Gemini-1.5-Flash}  & \multirow{1}{*}{Google}        & \textcolor{red}{\faTimesCircleO} & Candidate \\
    {Gemini-1.5-Pro}  & \multirow{1}{*}{Google}        & \textcolor{red}{\faTimesCircleO} & Examiner\&Candidate \\
    \hline
    {Claude-3.5-sonnet} & \multirow{2}{*}{Anthropic}     & \textcolor{red}{\faTimesCircleO} & Examiner\&Candidate \\
    {Claude-3-haiku}    &                                & \textcolor{red}{\faTimesCircleO} & Candidate \\
    \hline
    {GLM-4v-Plus}           & \multirow{1}{*}{Zhipu AI Inc.}   & \textcolor{red}{\faTimesCircleO} & Candidate \\
    \hline
    {Llama-3.2-90B-V}           & \multirow{1}{*}{Meta AI Inc.}   & \textcolor{green!75!black}{\faCheckCircleO} & Candidate \\
    \hline
    {Qwen2-VL-72B}          & \multirow{1}{*}{Alibaba}       & \textcolor{green!75!black}{\faCheckCircleO} & Candidate \\
    \bottomrule
    {Flux-1.1-Pro}          & \multirow{1}{*}{Black Forest Labs}       & \textcolor{red}{\faTimesCircleO} & Image generation \\
    \bottomrule
    \end{tabular}
\end{table*}

\begin{figure}[t]
    \centering
    \includegraphics[width=0.9\linewidth]{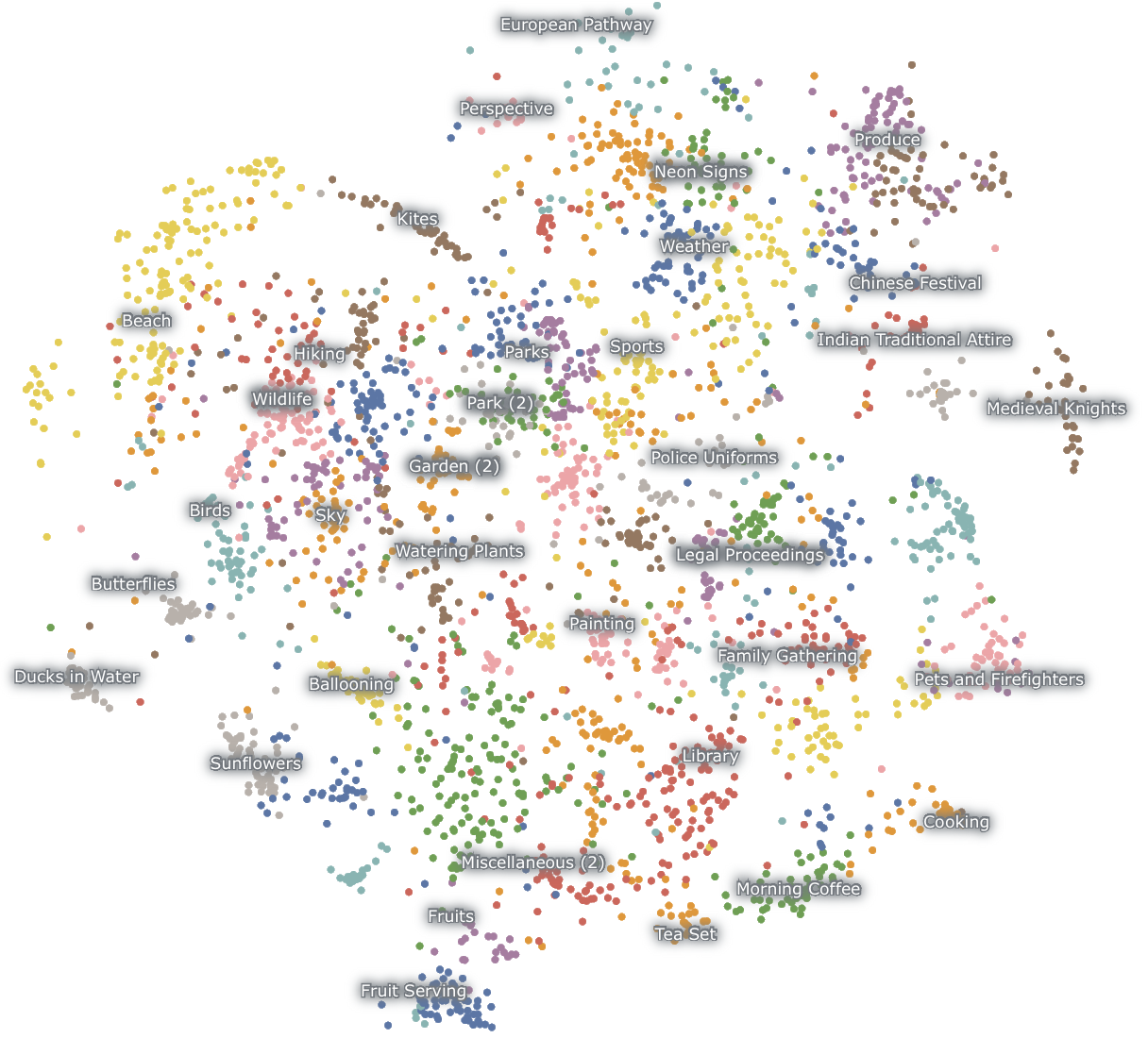}
    \captionsetup{justification=centerlast}
    \caption{Visualization of image descriptions in \textsc{AutoBench-V}.}
    \label{fig:all_image_description}
\end{figure}

\clearpage
\section{Details of User Input}
\label{sec:Appendix B}

In this section, we provide a comprehensive overview of the levels at which we categorize user inputs based on linguistic aspects. Our goal is to offer a comprehensive and broad representation of user requirements for LVLMs. However, as it is challenging to exhaustively cover every aspect, we base our categorization on aspects derived from the literature \citep{li2023seedbenchbenchmarkingmultimodalllms, meng2024mmiumultimodalmultiimageunderstanding, liu2023mmbenchmultimodalmodelallaround}. These aspects are considered representative and comprehensive examples of the capabilities of LVLMs and other aspects like in~\citep{chen2024we,xu2023lvlmehubcomprehensiveevaluationbenchmark} can be handled in a similar manner, without requiring additional fine-tuning or adjustments, as our framework is highly extensible, allowing users to propose their own aspects as needed.

\subsection{Basic Understanding} 

\textbf{Definition and Goal:} Basic Understanding refers to the recognition and identification of individual objects, characters, and scenes within an image. The goal is to accurately detect and label relevant elements, providing a foundation for more advanced tasks such as object tracking and scene interpretation~\citep{Wu2015ObjectTB, Xue2018ASO}.

\textbf{Requirement.} This task demands the ability to detect specific objects and differentiate between various types of objects. Additionally, it involves understanding the broader context of the scene and identifying real-life settings to enable accurate interpretation of the image's overall content.

\subsection{Spatial Understanding} 

\textbf{Definition and Goal.}
Spatial Understanding refers to the interpretation of the spatial arrangement and positioning of objects within an image~\citep{Cai2024SpatialBotPS, Guo2024RegionGPTTR}. The goal is to comprehend both two-dimensional and three-dimensional relationships, determining which objects are in the foreground or background, assessing their relative sizes and orientations, and understanding how they are positioned within the scene.

\textbf{Requirement.}
This task demands the ability to perceive depth, estimate distances between objects, and analyze how objects interact within the physical space of the image, providing a more accurate understanding of the spatial structure and context.

\subsection{Semantic Understanding} 

\textbf{Definition and Goal.}
Semantic Understanding involves interpreting the higher-level meaning and relationships within an image \citep{meng2024mmiumultimodalmultiimageunderstanding}. The goal is to move beyond simple object identification to understand the roles and interactions between objects, such as recognizing that a person is riding a bike or that two people are engaged in conversation. This level of understanding aims to capture the context and intent behind the scene, identifying how elements relate to each other to form a coherent narrative or message.

\textbf{Requirement.}
This task requires discerning the interactions and relationships between objects, understanding their roles within the scene, and interpreting the overall context to accurately derive the narrative or intended message conveyed by the image.

\subsection{Atmospheric Understanding} 

\textbf{Definition and Goal.}
Atmospheric Understanding focuses on grasping the mood, tone, and emotional ambiance conveyed by an image. The goal is to interpret not just what is depicted or how elements are arranged, but also how the scene feels and the emotional resonance it conveys to the viewer. For instance, an image of children laughing under warm sunlight in a lush park combines their expressions, bright colors, and soft lighting to create a joyful and carefree atmosphere.

\textbf{Requirement.}
This task requires the ability to capture and interpret subtle emotional cues and tonal qualities of the scene, distinguishing the overall mood and emotional impact of the image from more analytical aspects like semantic or spatial understanding.

\subsection{Reasoning Capacity} 

\textbf{Definition and Goal.}
Reasoning Capacity involves interpreting and analyzing the relationships and logical connections between different elements within an image~\citep{zhou2024navgpt, You2023IdealGPTID}. The goal is to infer potential outcomes, understand causal relationships, and make predictions about what might happen next based on visual cues. For example, if a person is holding an umbrella and the sky is dark, reasoning capacity would suggest that it might rain soon. This level also includes understanding abstract relationships, such as social dynamics or the intent behind actions, and making judgments about what is likely or possible given the visual information.

\textbf{Requirement.}
This task requires the ability to analyze logical connections between elements, infer outcomes, and understand causal relationships, as well as to interpret abstract concepts and make predictions based on the visual context.

\section{Details of Difficulty Grading}
\label{Appendix C}
This section describes in detail the difficulty levels for the pictures (\autoref{sec:image}) and questions (\autoref{sec:question}) used in prompts respectively. The following is the instruction guiding the examiner model to generate image descriptions and questions of varying difficulty levels. For details, see \autoref{prompt:description generation}. Moreover, the difficulty levels are defined manually and are not fixed, allowing for flexible adjustments based on the user’s specific requirements.

\subsection{Image Description}
\label{sec:image}
\textbf{Easy difficulty.} 
Focus on a single subject but with minor additional complexity to test precision in details or context. Establish baseline capability with subtle challenges, such as fine texture, simple interactions, or slight variations in lighting. A single object or entity with some basic contextual details or additional features, placed against a minimally distracting background. Test finer details (e.g., texture, reflection) while still keeping the composition simple and clear. For examples: "A perfectly polished red apple with a small leaf attached, placed on a slightly reflective white surface."

\textbf{Medium difficulty.} 
Scenes involve multiple elements or interactive settings that require nuanced spatial arrangement and accurate relationships between objects. Evaluate the ability to generate cohesive, moderately dynamic scenes with layered realism and a stronger sense of depth. Description Requirements: Include multiple objects interacting naturally in a believable environment, with more intricate details and subtle light or shadow effects. Incorporate realistic environmental elements and ensure spatial coherence, emphasizing interactions and secondary details (e.g., shadows, water splashes). For examples: "A golden retriever running on a sandy beach, splashing water as it chases a bright orange ball, with distant waves and a partly cloudy sky."

\textbf{Hard difficulty.} 
Scenes incorporate high complexity with multiple interdependent elements, challenging perspectives, or dynamic and intricate lighting or textural effects. Push the limits of rendering capability to handle advanced relationships, environmental effects, and challenging compositions. The scene must include 3-4 main elements or a combination of dynamic features, such as motion, light interplay, or atmospheric conditions, while maintaining clarity and logic. Highlight challenges such as complex reflections, dynamic light, or multi-element interactions, ensuring visual harmony and detailed textures. For examples: "A raindrop-streaked window reflecting the interior of a cozy room, with a black cat sitting on the windowsill and a glowing city skyline visible through the glass, all under the warm hues of a sunset."

\subsection{Question}
\label{sec:question}
\textbf{Easy difficulty.} Focus on questions that require identifying simple, prominent, and explicit details within the image. These questions should be straightforward, relying solely on basic observation without the need for inference or interpretation. For example, you might ask about the color of a specific object, the presence of a single item, or the shape of an easily recognizable feature. The key is to keep the questions direct and simple, ensuring that the answer is obvious and immediately visible in the image.
    
\textbf{Medium difficulty.} Design questions that necessitate a moderate level of observation and inference. These questions should involve understanding relationships between elements, recognizing interactions, or identifying less prominent features that are still clear but not immediately obvious. Examples could include questions about the relative position of objects, identifying an action taking place, or understanding the context of a scene. The goal is to require some level of thought beyond basic observation, challenging the model to understand the scene's composition or narrative without being overly complex.

\textbf{Hard difficulty.} Create questions that require the model to notice and interpret more detailed aspects of the image. These questions should involve recognizing multiple elements working together, understanding more complex interactions, or identifying details that are present but not immediately obvious. For example, you might ask about the positioning of objects relative to each other in a more crowded scene, subtle changes in lighting or color that affect the appearance of objects, or identifying an element that is not the main focus but is still visible in the background. The aim is to challenge the model to go beyond surface-level details, but without making the task too abstract or overly difficult.

\section{Human Evaluation}
\label{humaneval detail}

\subsection{Details of Human Evaluation}
\begin{figure*}[ht]
    \centering
    \includegraphics[width=0.8\linewidth]{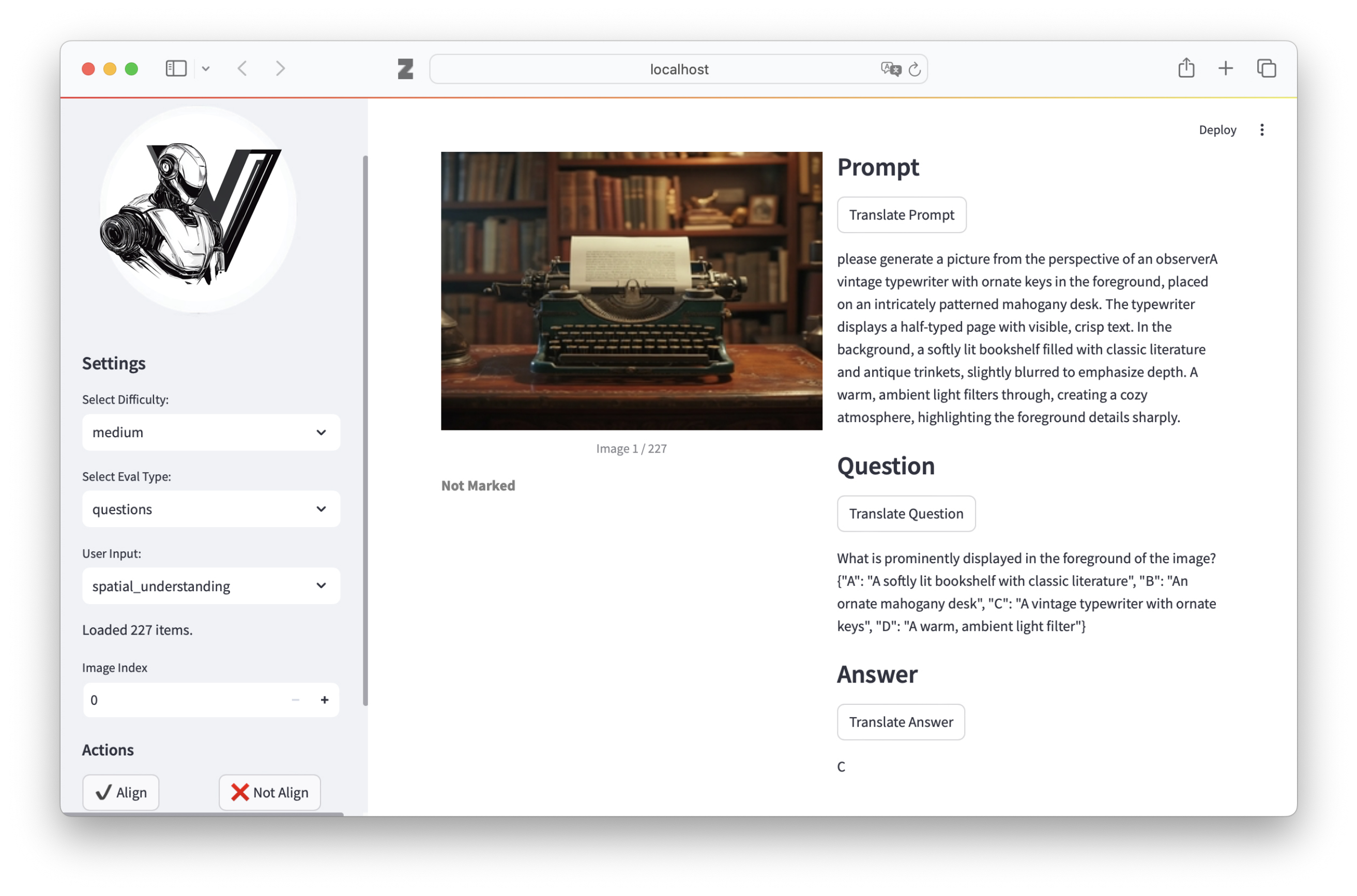}
    \captionsetup{justification=centerlast}
    \caption{Screenshot of Human Evaluation (Example 1).}
    \label{fig:screenshot1}
\end{figure*}

\begin{figure*}[ht]
    \centering
    \includegraphics[width=0.8\linewidth]{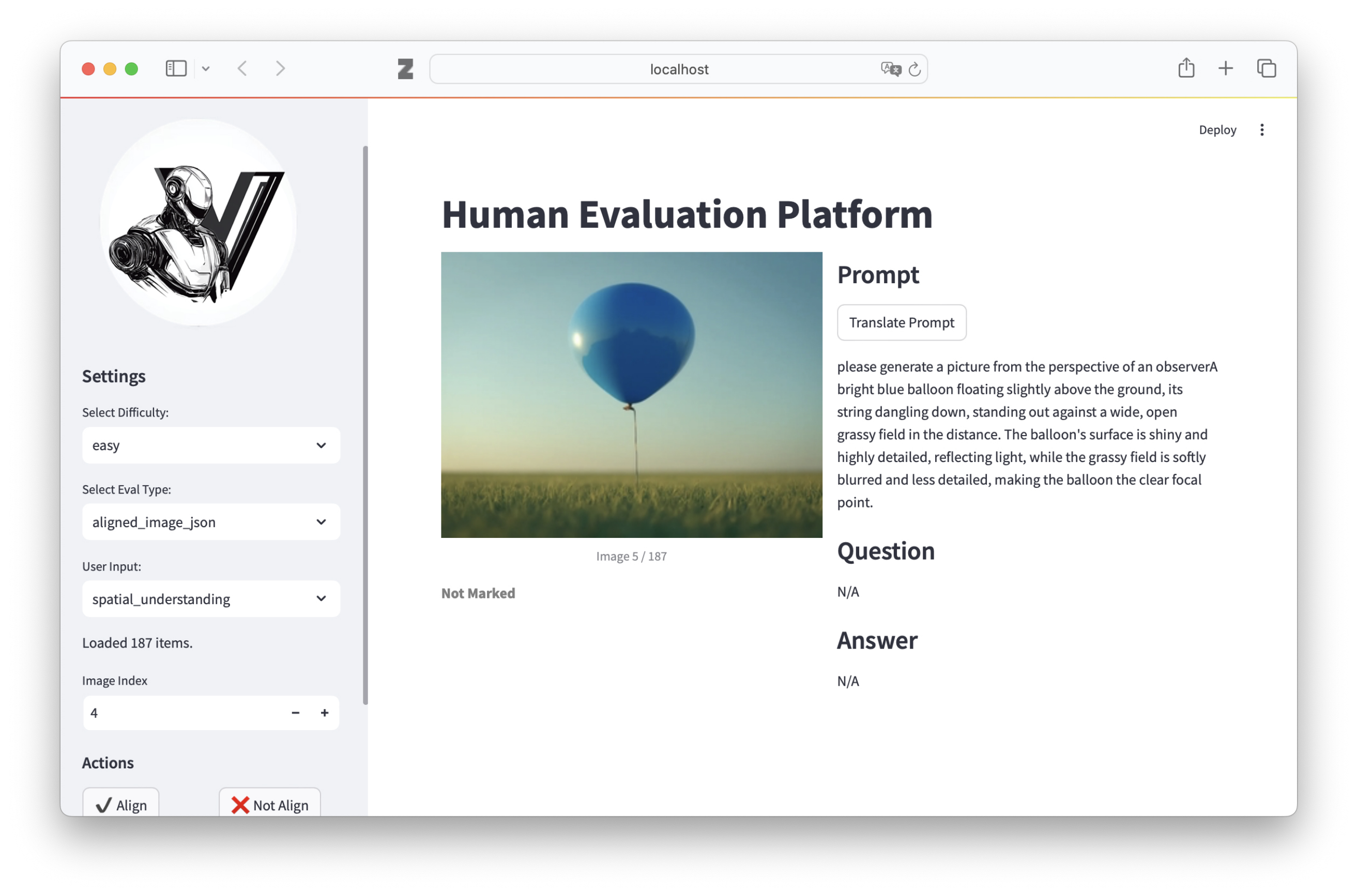}
    \captionsetup{justification=centerlast}
    \caption{Screenshot of Human Evaluation (Example 2).}
    \label{fig:screenshot2}
\end{figure*}

The evaluation was carried out by a panel of five evaluators: three undergraduate students and two PhD students, all possessing professional English skills. Sample annotation screenshots from the human evaluation process are presented in \autoref{fig:screenshot1} and \autoref{fig:screenshot2}. To ensure unbiased results, each evaluator independently assessed all samples. A sample was considered aligned if it received a majority vote (\emph{i.e.}, more than half of the evaluators agreed on its alignment).

\subsection{Human Evaluation Guidelines}
In this section, we outline the guidelines followed during human evaluations to ensure reliability and validity.

For \textbf{Description Generation Guideline}, the evaluators need to consider the following three points:

$\triangleright$ \textbf{Alignment with Image:} The main criterion is how well the generated description reflects the visual content. Descriptions must accurately correspond to the image, avoiding vague or abstract expressions, as well as hallucination \citep{huang2024position}. Each description should provide clear, specific details that align with the image content and the defined fine-grained aspects.

$\triangleright$ \textbf{Specificity and Clarity:} Ensure that descriptions are specific, directly related to the image, and free from ambiguous or overly generalized language.

$\triangleright$ \textbf{Relevance to Aspects:} Assess whether the description aligns with the corresponding themes and expected content. Descriptions must clearly communicate the intended visual elements and avoid any misalignment between the image and the description.

For \textbf{Question-Answer Alignment}, there are two points that the evaluators should consider inspired by the previous study \citep{liu2023alignbench}:

$\triangleright$ \textbf{Clarity and Accuracy:} Each question must be clear, unambiguous, and directly derived from the image. The answers should correspond to observable details or logical inferences from the image, with only one correct answer for each question. There should be no irrelevant or misleading information in the questions or answers.

$\triangleright$ \textbf{Consistency with Image:} Verify that both the question and answer are directly based on the image's content. The evaluation should ensure that there is a logical and clear relationship between the visual cues and the generated question-answer pair, particularly for tasks involving higher difficulty.

\clearpage
\section{Notations}
\label{app:notation}
\begin{table}[ht]
\centering
\small
\renewcommand{\arraystretch}{1.2}
\captionsetup{justification=centerlast}
\caption{Notation used in the methodology description.}
\label{tab:notation}
\scalebox{0.8}{
\begin{tabular}{ll}
\toprule[1pt]
\textbf{Notation} & \textbf{Description} \\
\midrule
\(q\) & User input specifying the evaluation target or focus. \\
\(n\) & Number of \textit{general aspects}. \\
\(\{A^{(g)}_1, A^{(g)}_2, \dots, A^{(g)}_n\}\) & Set of \(n\) general aspects, each representing a high-level capability dimension. \\
\(A^{(g)}_i\) & The \(i\)-th general aspect. \\
\(m\) & Number of \textit{fine-grained aspects} under each general aspect. \\
\(\{A^{(f)}_{i1}, A^{(f)}_{i2}, \dots, A^{(f)}_{im}\}\) & Set of \(m\) fine-grained aspects under the \(i\)-th general aspect. \\
\(A^{(f)}_{ij}\) & The \(j\)-th fine-grained aspect under the \(i\)-th general aspect. \\
\(\mathcal{M}_v\) & The LVLM used for aspect generation, description generation, and self-validation. \\
\(\mathcal{A}\) & The complete set of generated aspects, i.e., \(\bigcup_{i=1}^n \{A^{(g)}_i\} \cup \{A^{(f)}_{ij}\}\). \\
\(|\mathcal{A}|\) & The cardinality of \(\mathcal{A}\), with \(|\mathcal{A}| = m \times n\). \\
\midrule
\(d\in\{\text{easy}, \text{medium}, \text{hard}\}\) & Difficulty level for image description generation. \\
\(\omega\) & Number of image descriptions generated per aspect at each difficulty level. \\
\(\{\mathcal{T}_{ij1}^d, \dots, \mathcal{T}_{ij\omega}^d\}\) & Set of \(\omega\) image descriptions for the fine-grained aspect \(A^{(f)}_{ij}\) at difficulty \(d\). \\
\(\mathcal{T}_{ij}^d\) & A specific image description for aspect \(A^{(f)}_{ij}\) at difficulty \(d\). \\
\midrule
\(t_e\) & Topic word selected at iteration \(e\). \\
\(K_e = \{k_{e1}, k_{e2}, \dots, k_{ec}\}\) & Set of \(|c|\) keywords related to the topic word \(t_e\) at iteration \(e\). \\
\(G_e=(V_e, E_e)\) & Semantic graph at iteration \(e\), whose nodes are topic/keyword sets and edges are semantic relationships. \\
\(S_e\) & Combination of \(V_{e-1}\), \(t_e\), and \(K_e\) before applying the exclusion mechanism. \\
\(V_e\) & Refined node set (topic words/keywords) after excluding high-degree nodes in iteration \(e\). \\
\(f(S_e)\) & A function that determines how many top-degree nodes to exclude from \(S_e\). \\
\midrule
\(\{\mathcal{I}_{ij}^d\}\) & Generated images corresponding to the image description(s) \(\mathcal{T}_{ij}^d\). \\
\(\Phi_{ij}^d = \{\phi_{ij1}^d, \dots, \phi_{ijp}^d\}\) & Set of \(p\) simple VQA-style questions for self-validation of \(\mathcal{I}_{ij}^d\). \\
\(p\) & Number of questions for each image’s self-validation. \\
\(\mathcal{F}(\mathcal{I}_{ij}^d, \mathcal{T}_{ij}^d, \Phi_{ij}^d)\) & Self-validation function returning the alignment score \(S_{ij}^d\). \\
\(S_{ij}^d\) & Alignment score, i.e., the proportion of correctly answered questions among \(\Phi_{ij}^d\). \\
\(\zeta\) & Threshold for deciding whether to regenerate an image or keep it with recorded errors. \\
\(\mathcal{E}_{ij}^d\) & Noted error(s) if the image partially misaligns with the description (\(S_{ij}^d < 1\)). \\
\bottomrule[1pt]
\end{tabular}}
\end{table}

\clearpage
\section{Error Study}

\begin{figure}[ht]
    \centering
    \includegraphics[width=0.7\linewidth]{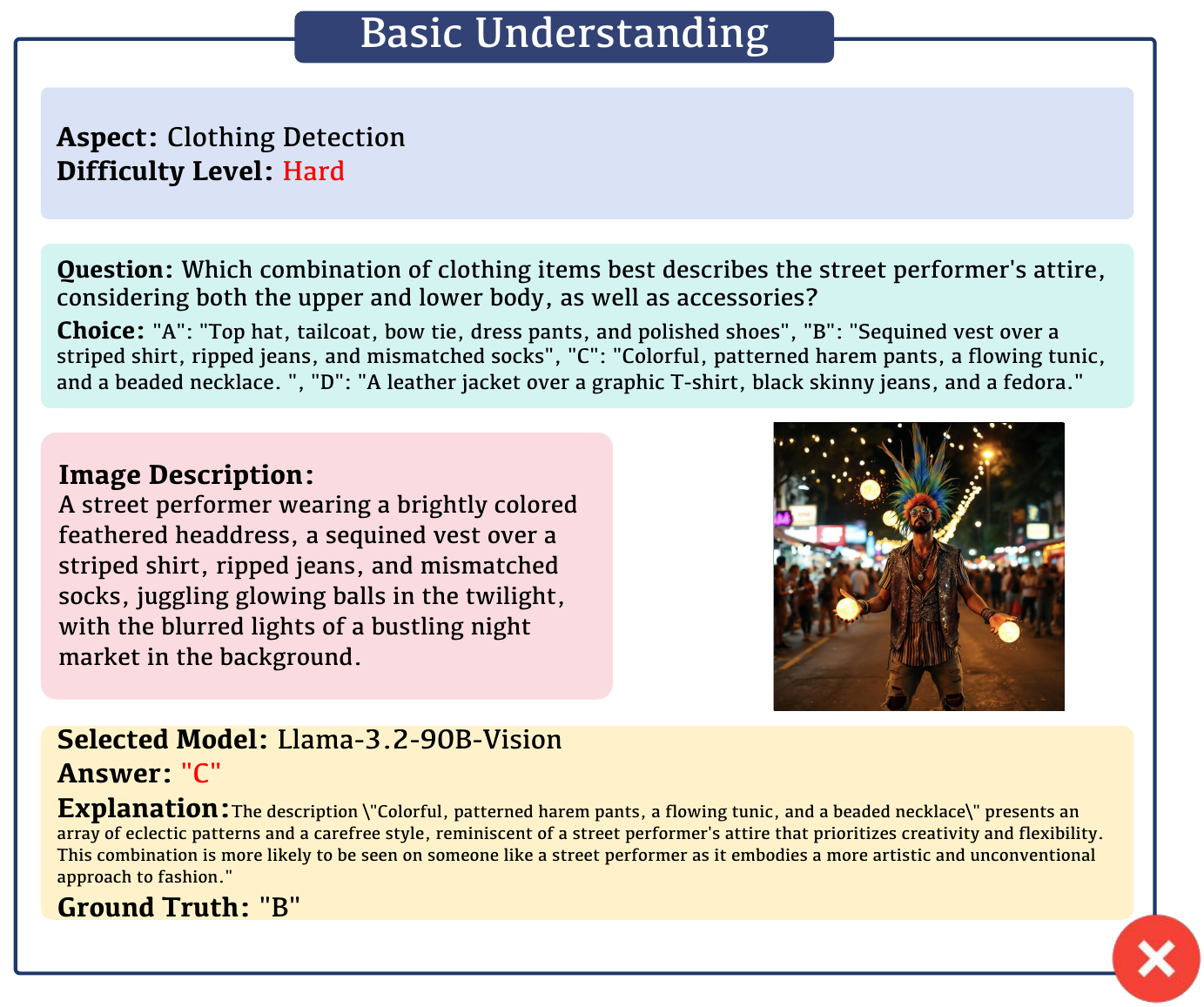}
    \captionsetup{justification=centerlast}
    \caption{Error study of Llama-3.2-90B-Vision under basic understanding.}
    \label{fig:basic_error}
\end{figure}

\begin{figure}[ht]
    \centering
    \includegraphics[width=0.7\linewidth]{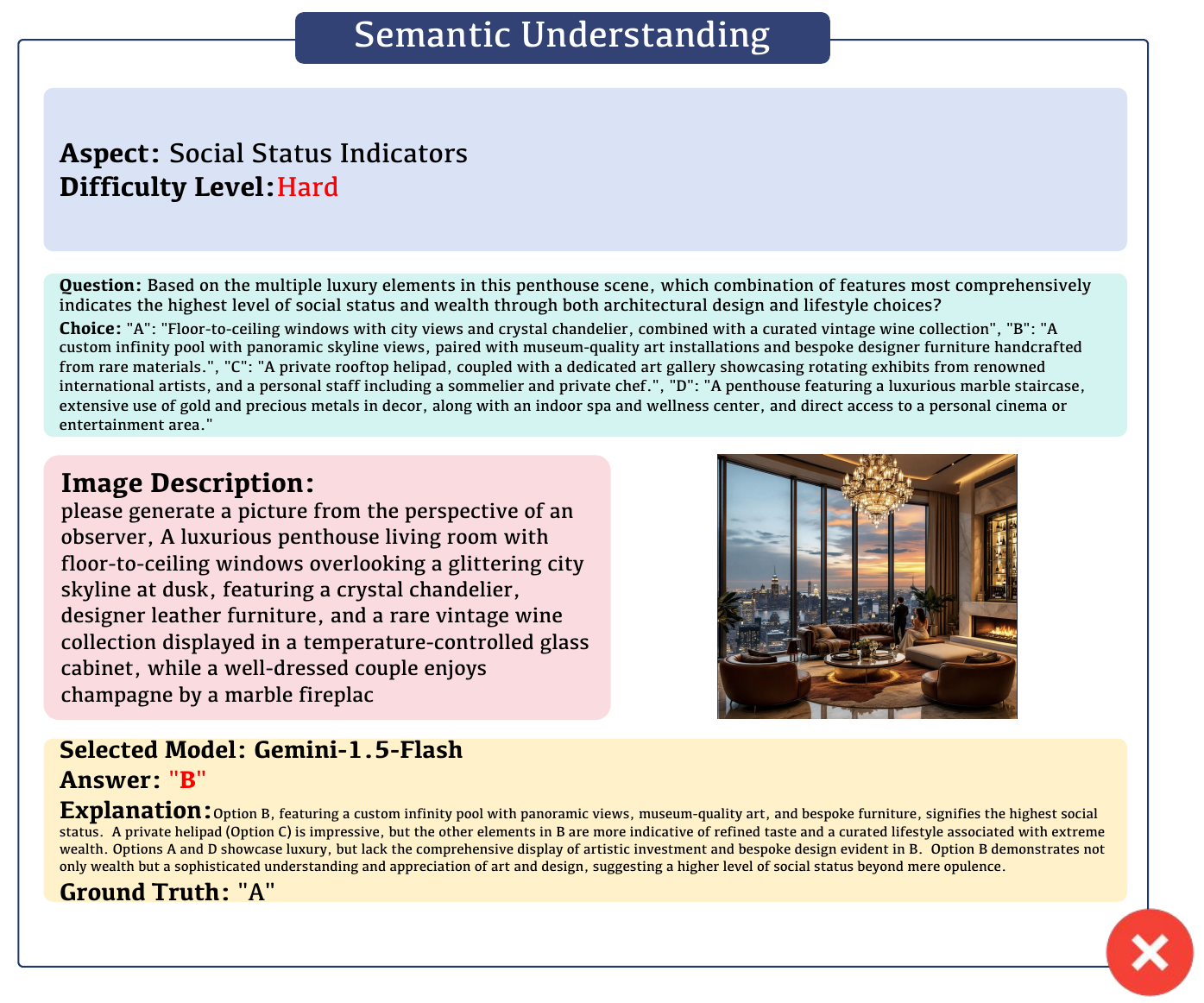}
    \captionsetup{justification=centerlast}
    \caption{Error study of Gemini-1.5-Flash under semantic understanding.}
    \label{fig:semantic_error}
\end{figure}

\begin{figure}[ht]
    \centering
    \includegraphics[width=0.7\linewidth]{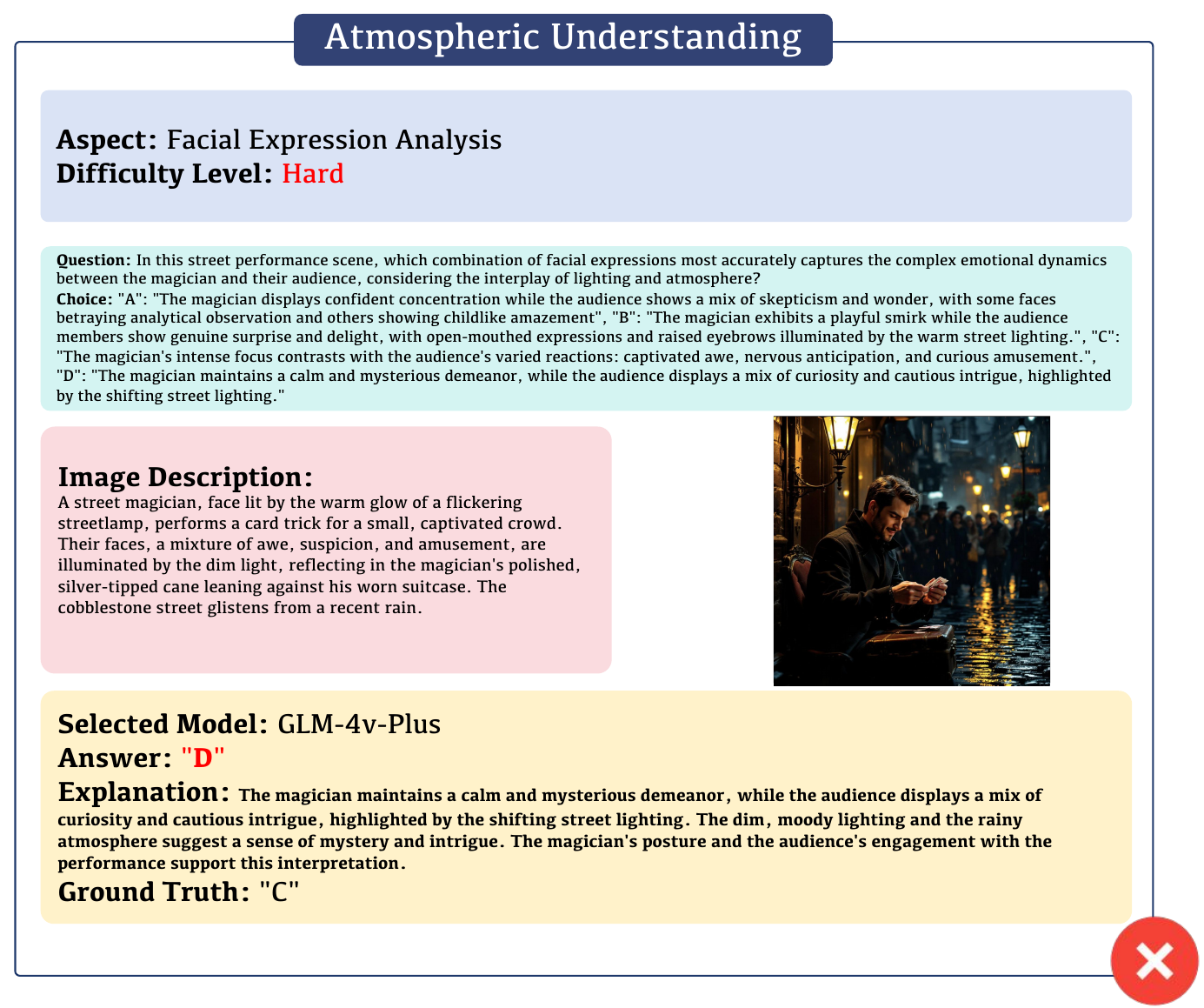}
    \captionsetup{justification=centerlast}
    \caption{Error study of GLM-4v-Plus under atmospheric understanding.}
    \label{fig:atmospheric_error}
\end{figure}

Through extensive experimental analysis, we have categorized the common problems encountered by LVLMs in VQA tasks into two main types: \textit{image comprehension errors} and \textit{image reasoning errors}. Regarding the first category, LVLMs often fail to truly understand the details in an image. For instance, in \autoref{fig:basic_error}, the model failed to accurately identify the \textit{performer’s attire}. In \autoref{fig:semantic_error}, from the model’s explanation, it is evident that the focus of its understanding was placed on the text rather than the image. Option B mentions \textit{a swimming pool}, option C references \textit{a helipad}, and option D refers to \textit{a health center}—none of which are present in the image. In \autoref{fig:atmospheric_error}, the model failed to correctly identify the \textit{magician’s facial expression} as a smile. Therefore, it cannot be associated with calmness or scheming; instead, it represents a confident smile. These errors demonstrate the model's inadequate comprehension of image details.

\begin{figure}[ht]
    \centering
    \includegraphics[width=0.7\linewidth]{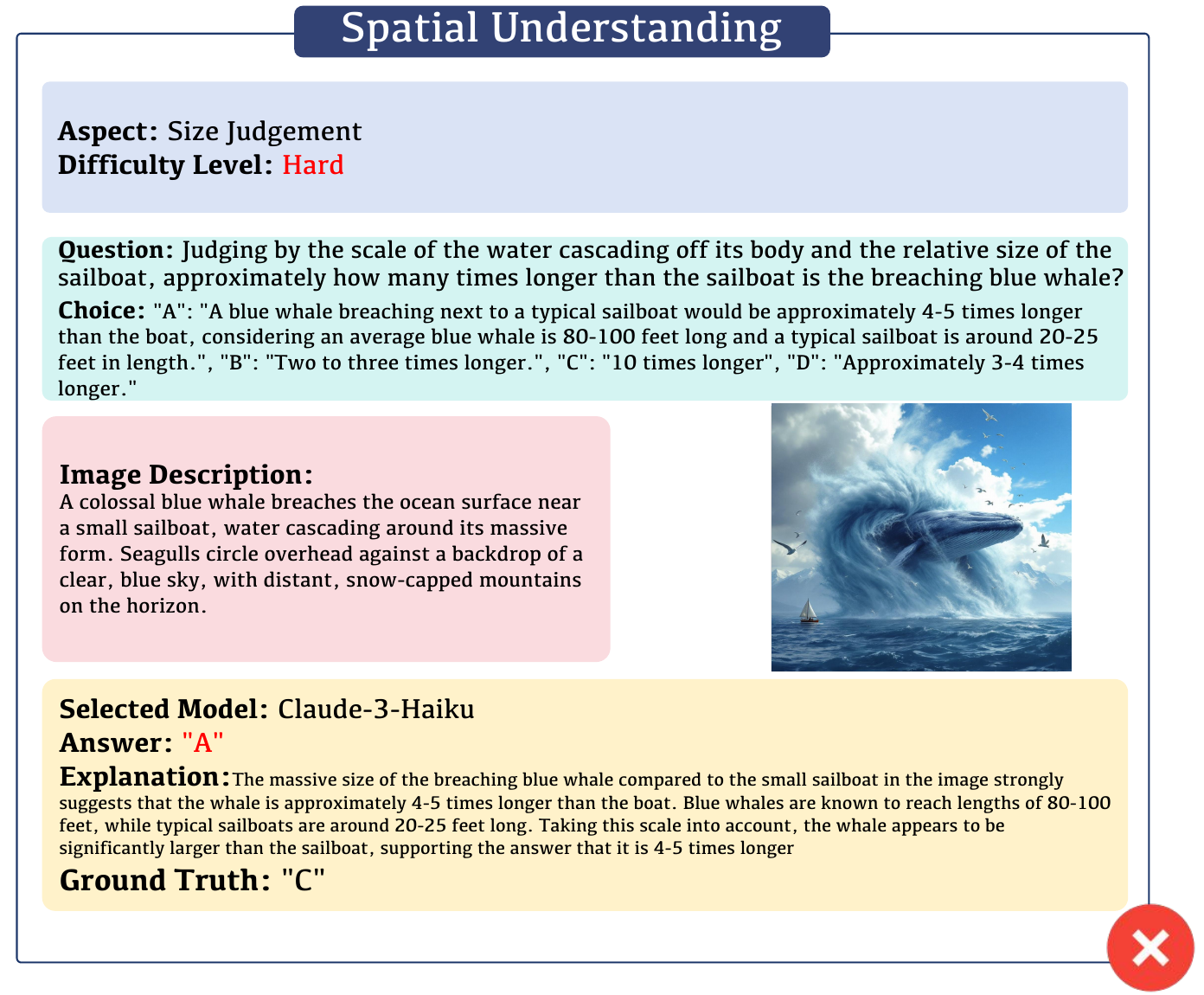}
    \captionsetup{justification=centerlast}
    \caption{Error study of Claude-3-Haiku under spatial understanding.}
    \label{fig:spatial_error}
\end{figure}

\begin{figure}[ht]
    \centering
    \includegraphics[width=0.7\linewidth]{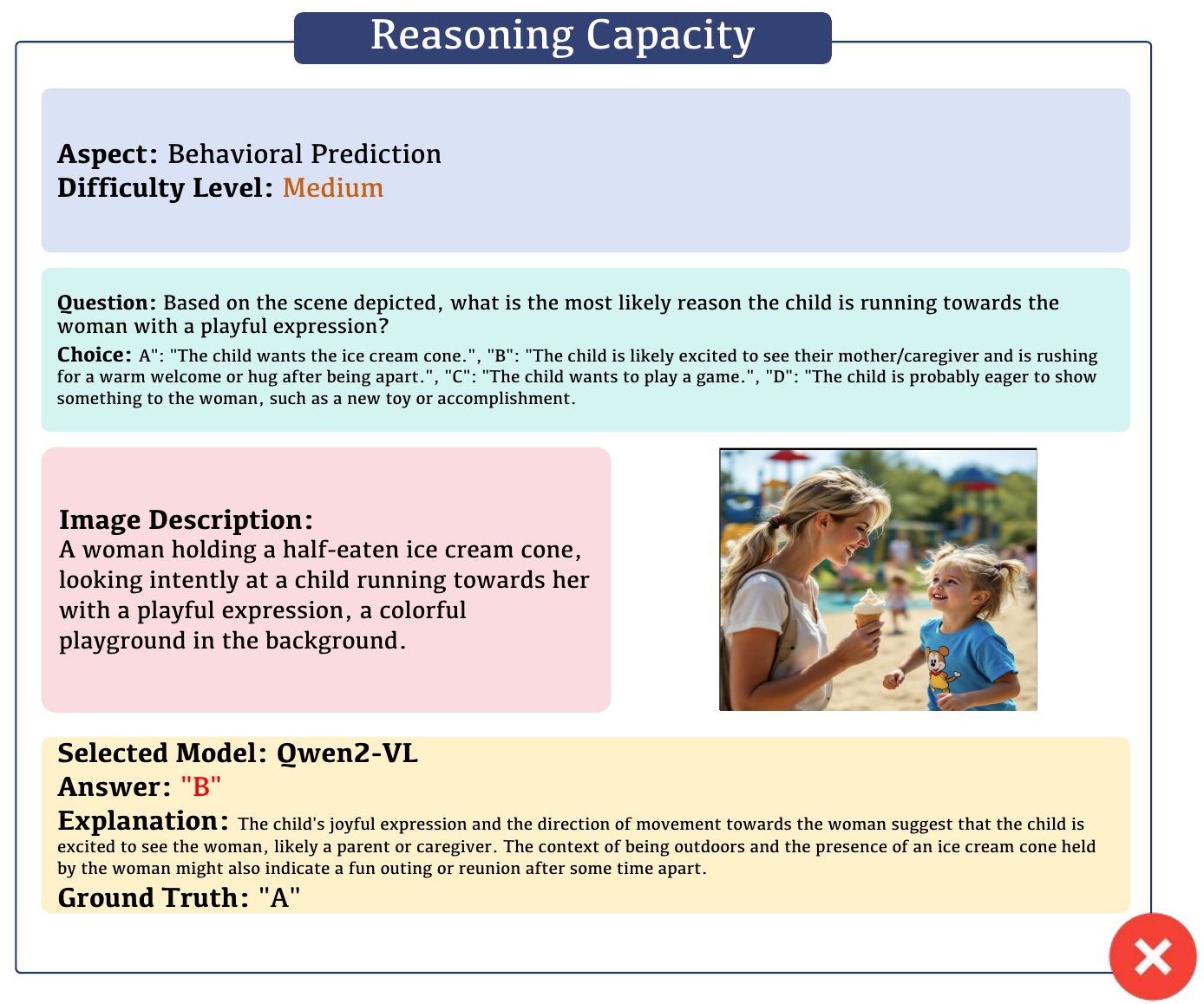}
    \captionsetup{justification=centerlast}
    \caption{Error study of Qwen2-VL under reasoning capacity.}
    \label{fig:reasoning_error}
\end{figure}

Image reasoning errors occur when models accurately perceive the image content but falter in their reasoning process, leading to incorrect answers. For instance, in \autoref{fig:spatial_error}, the model correctly identified the \textit{blue whale} and \textit{the small boat} in the image. However, based on its explanation, it relied more on its prior knowledge rather than the image itself, leading to an incorrect answer. In \autoref{fig:reasoning_error}, model accurately recognized the excitement and joy on the little girl’s face but misinterpreted the intent. From the image, it is clear that the woman is holding a complete ice cream cone and is about to hand it to the girl, indicating that the girl desires the ice cream.

\clearpage
\section{Case Study}

\begin{figure}[ht]
    \centering
    \includegraphics[width=0.7\linewidth]{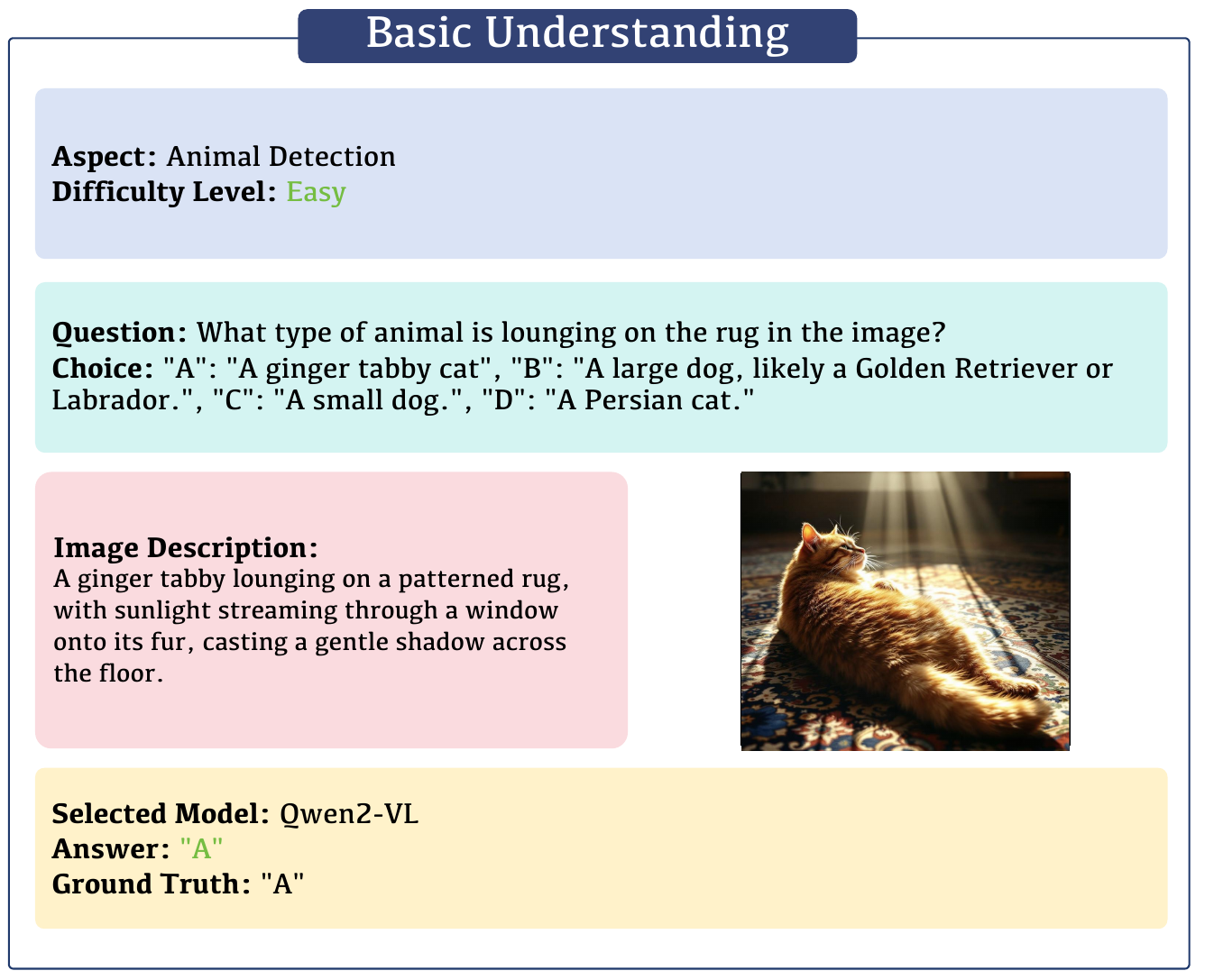}
    \captionsetup{justification=centerlast}
    \caption{Case Study 1.}
    \label{fig:case_1}
\end{figure}

\begin{figure}[ht]
    \centering
    \includegraphics[width=0.7\linewidth]{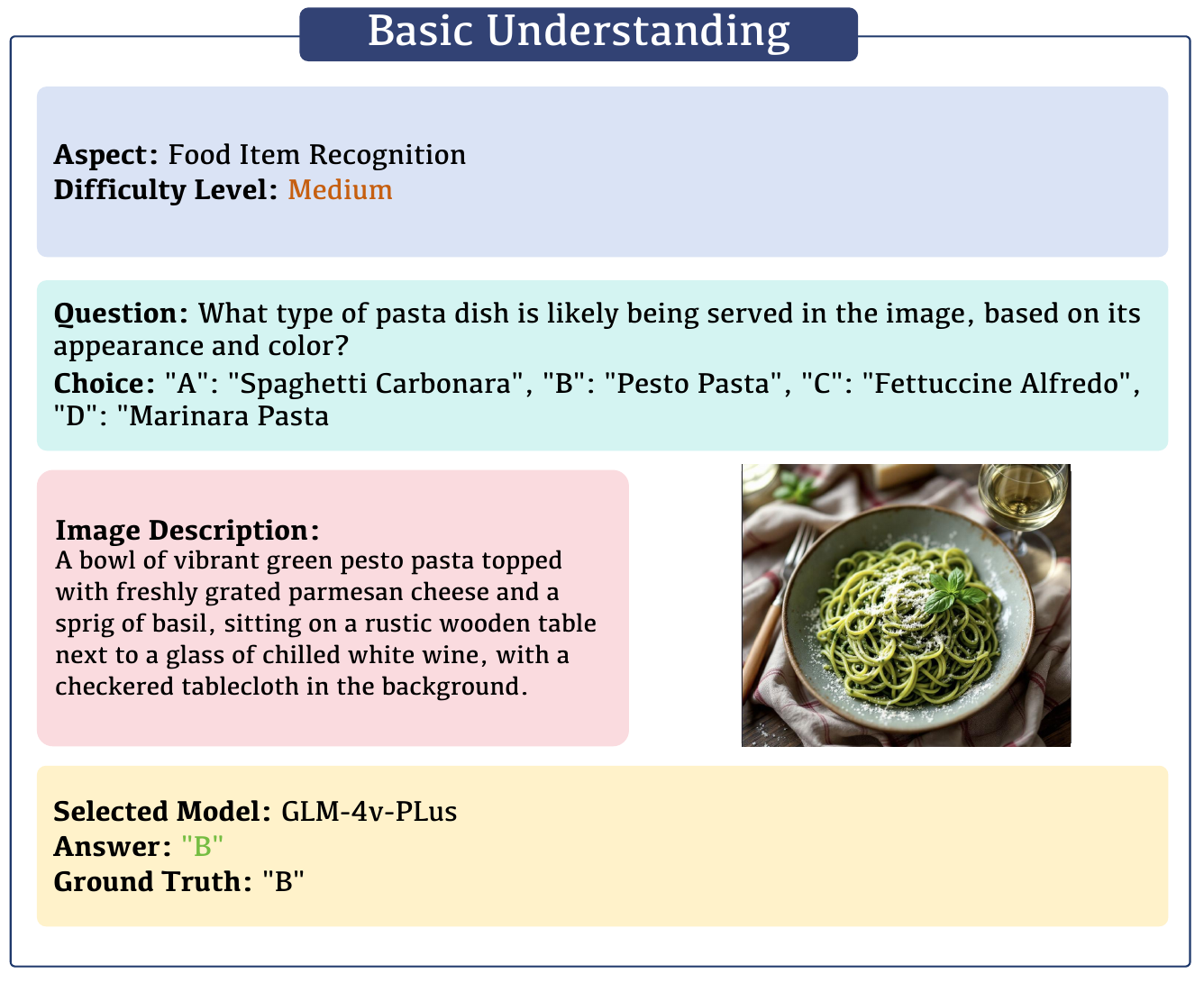}
    \captionsetup{justification=centerlast}
    \caption{Case Study 2.}
    \label{fig:case_2}
\end{figure}

\begin{figure}[ht]
    \centering
    \includegraphics[width=0.7\linewidth]{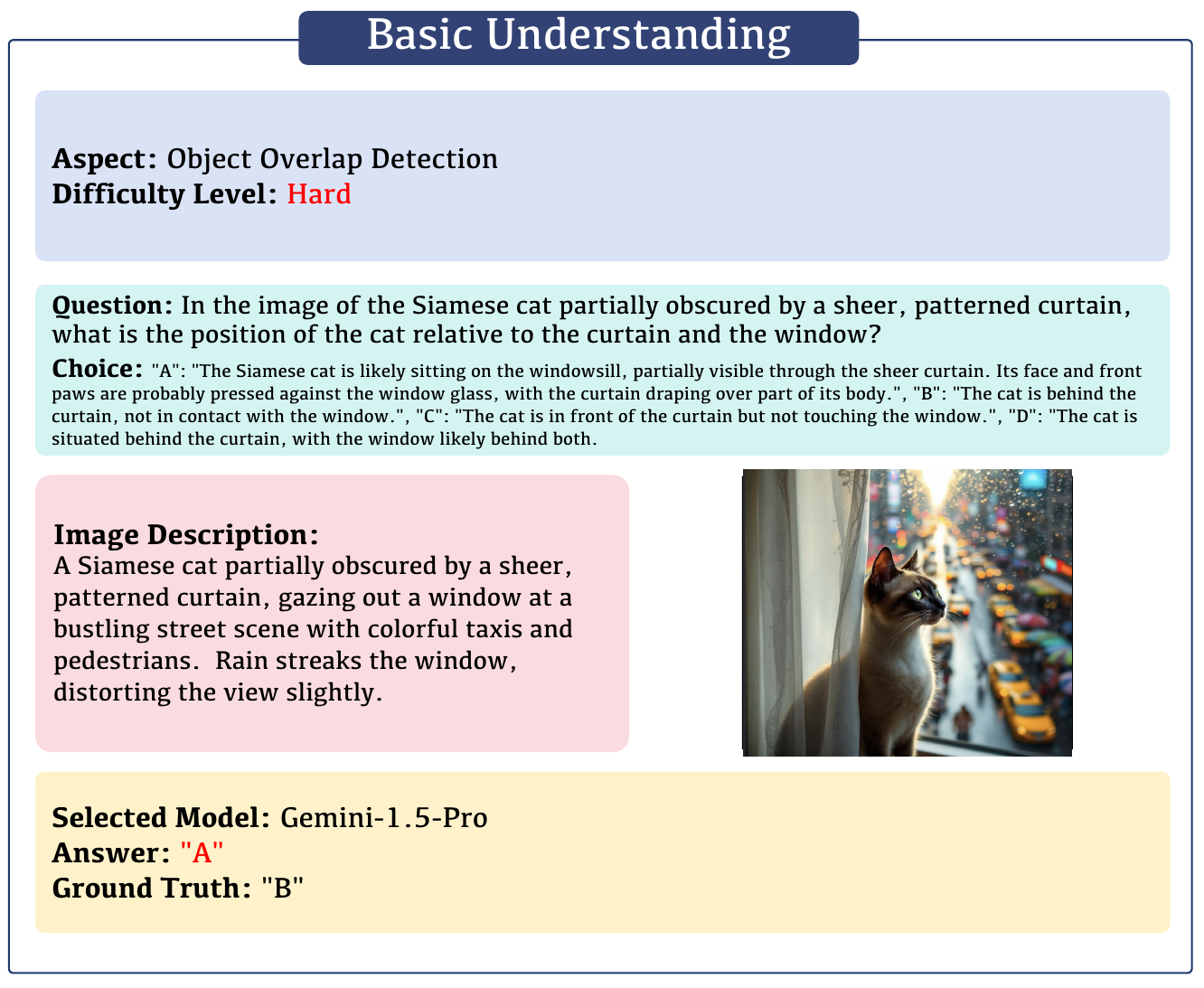}
    \captionsetup{justification=centerlast}
    \caption{Case Study 3.}
    \label{fig:case_3}
\end{figure}

\begin{figure}[ht]
    \centering
    \includegraphics[width=0.7\linewidth]{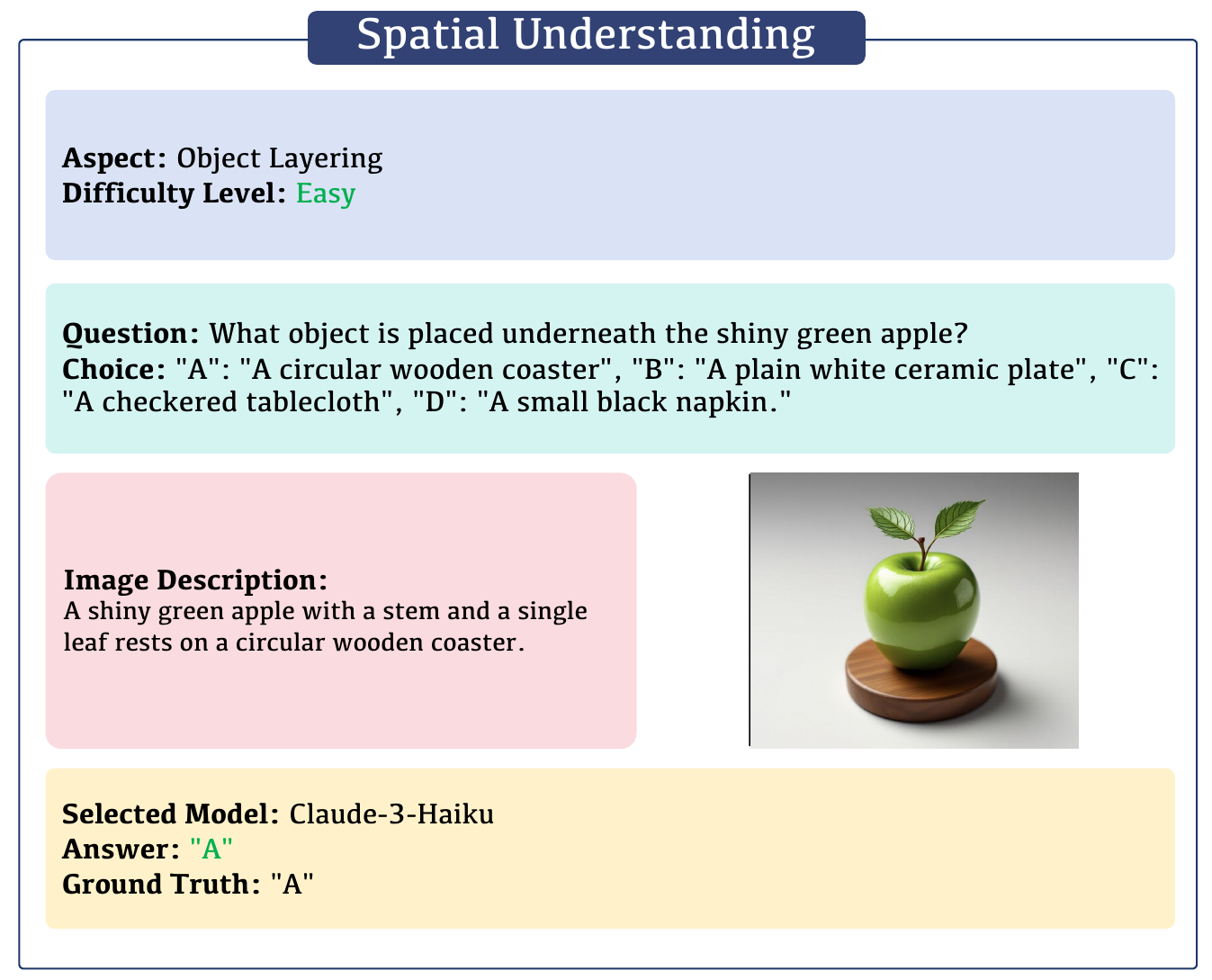}
    \captionsetup{justification=centerlast}
    \caption{Case Study 4.}
    \label{fig:case_4}
\end{figure}

\begin{figure}[ht]
    \centering
    \includegraphics[width=0.7\linewidth]{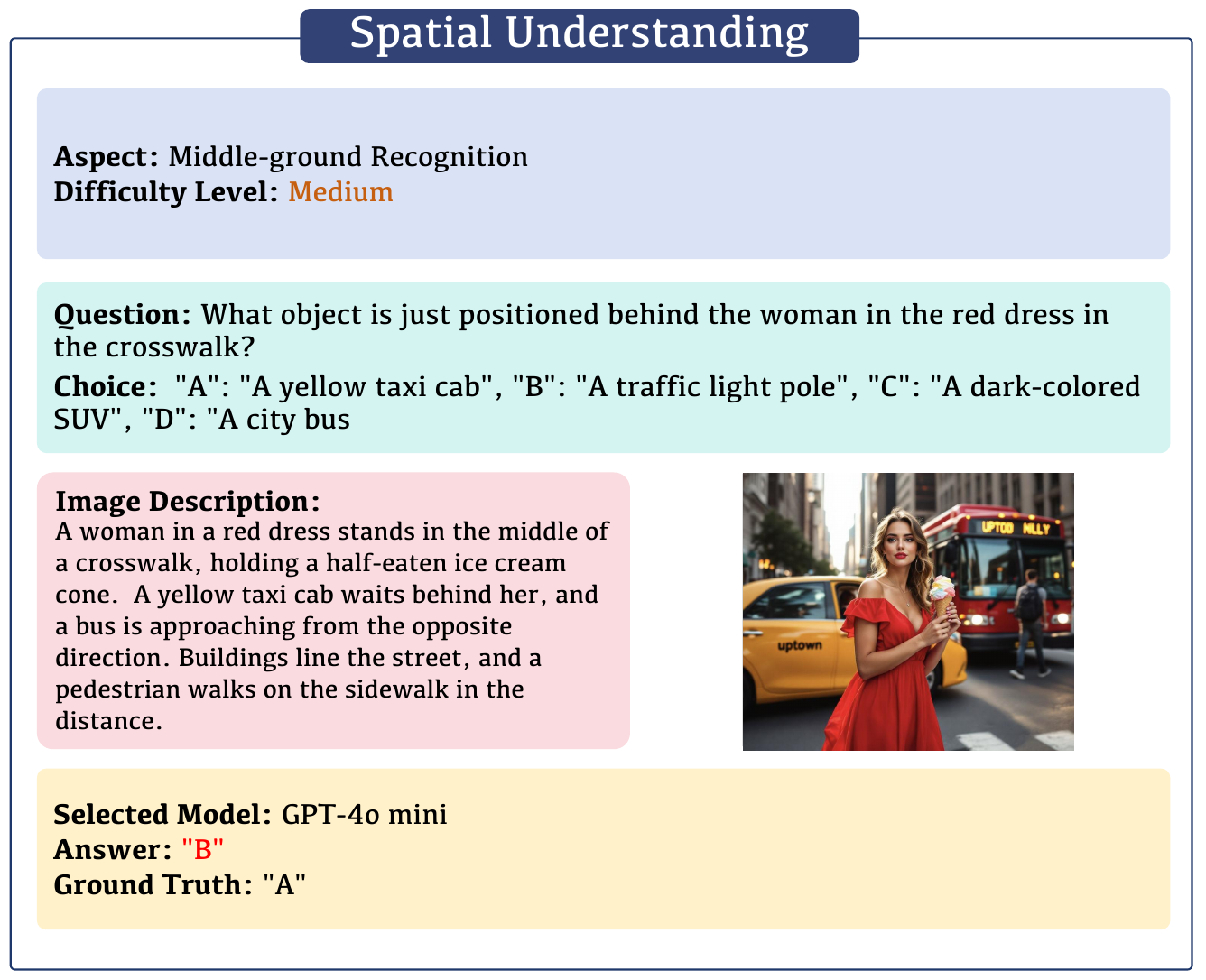}
    \captionsetup{justification=centerlast}
    \caption{Case Study 5.}
    \label{fig:case_5}
\end{figure}


\begin{figure}[ht]
    \centering
    \includegraphics[width=0.7\linewidth]{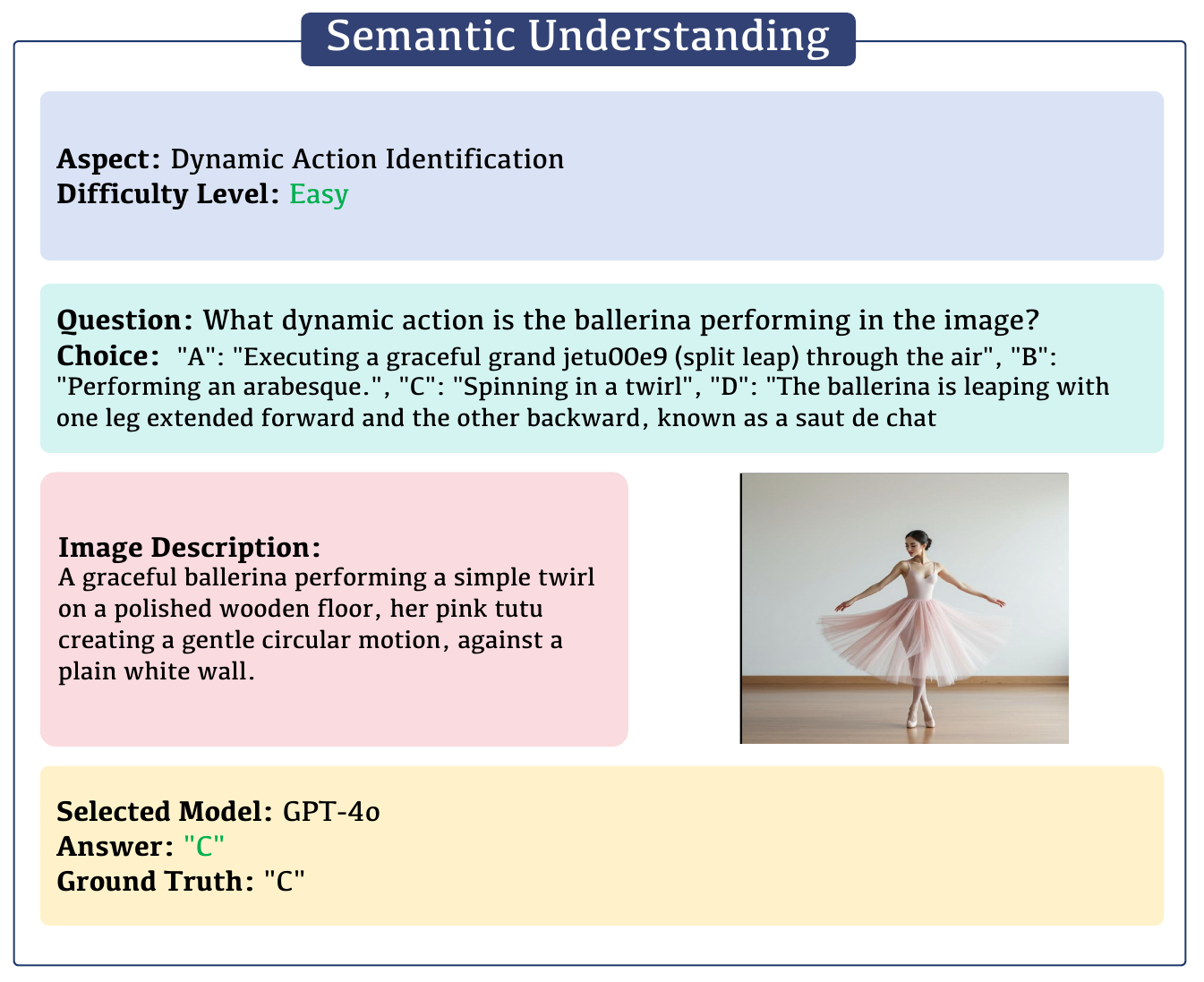}
    \captionsetup{justification=centerlast}
    \caption{Case Study 6.}
    \label{fig:case_6}
\end{figure}

\begin{figure}[ht]
    \centering
    \includegraphics[width=0.7\linewidth]{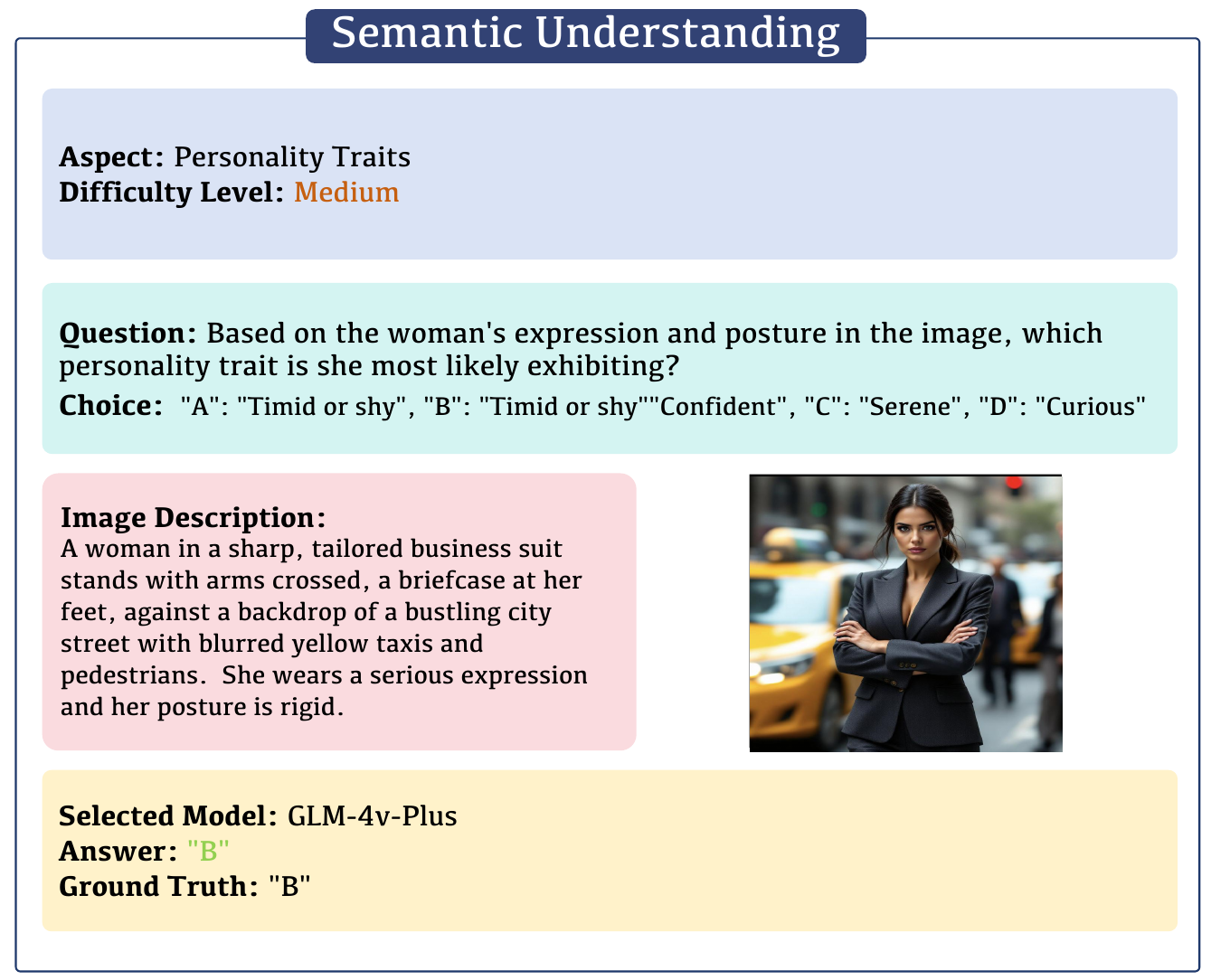}
    \captionsetup{justification=centerlast}
    \caption{Case Study 7.}
    \label{fig:case_7}
\end{figure}

\begin{figure}[ht]
    \centering
    \includegraphics[width=0.7\linewidth]{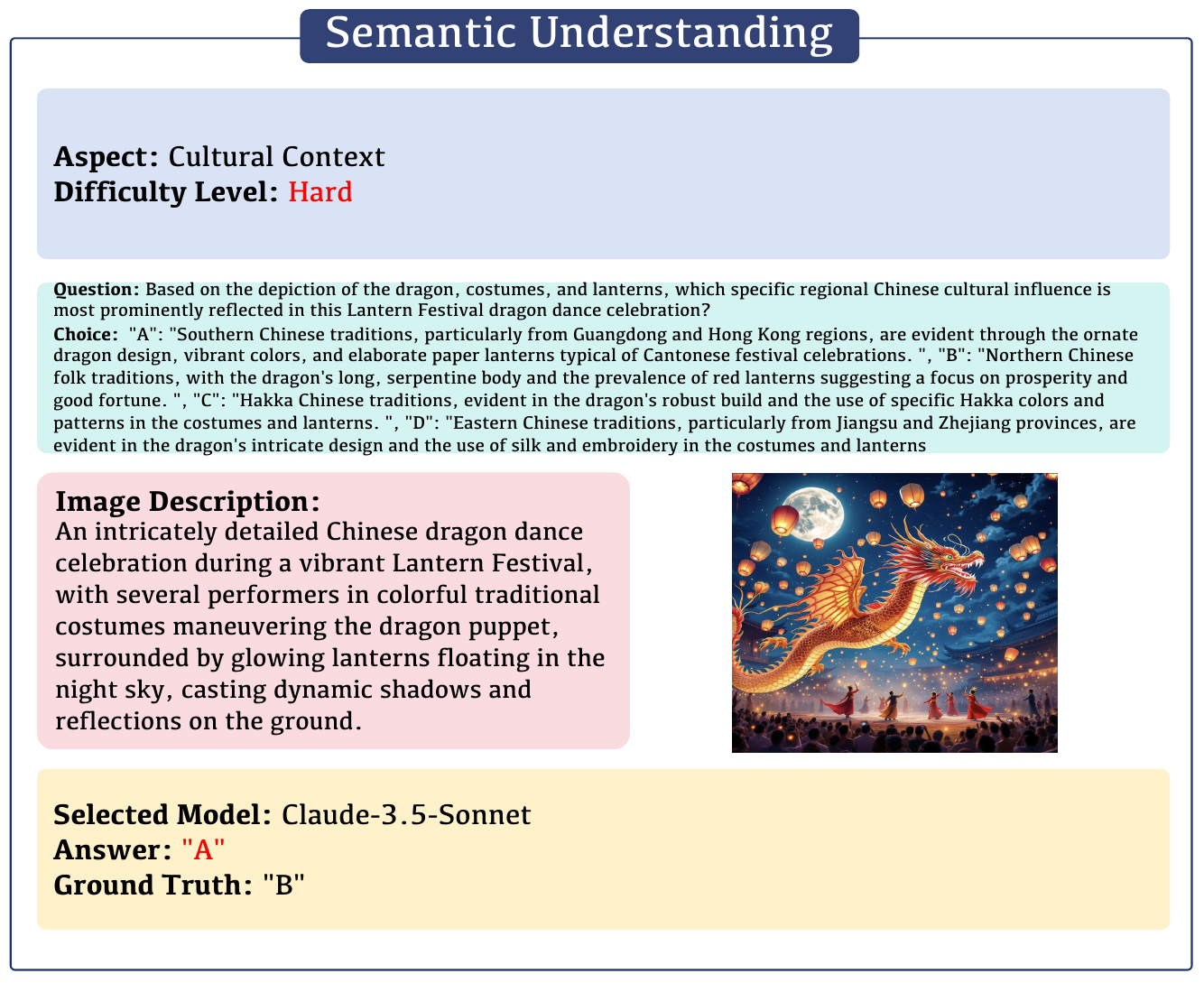}
    \captionsetup{justification=centerlast}
    \caption{Case Study 8.}
    \label{fig:case_8}
\end{figure}

\begin{figure}[ht]
    \centering
    \includegraphics[width=0.7\linewidth]{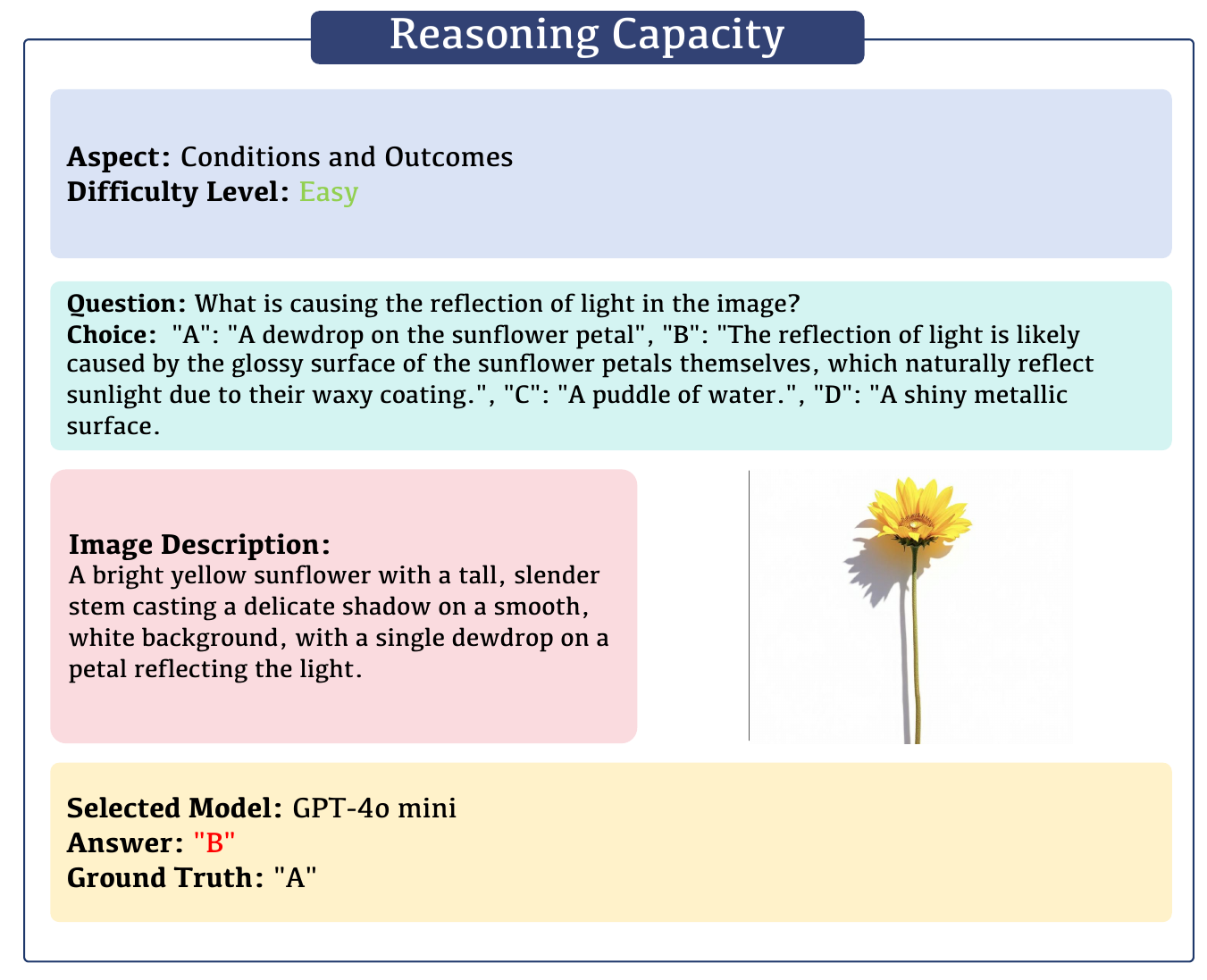}
    \captionsetup{justification=centerlast}
    \caption{Case Study 9.}
    \label{fig:case_9}
\end{figure}

\begin{figure}[ht]
    \centering
    \includegraphics[width=0.7\linewidth]{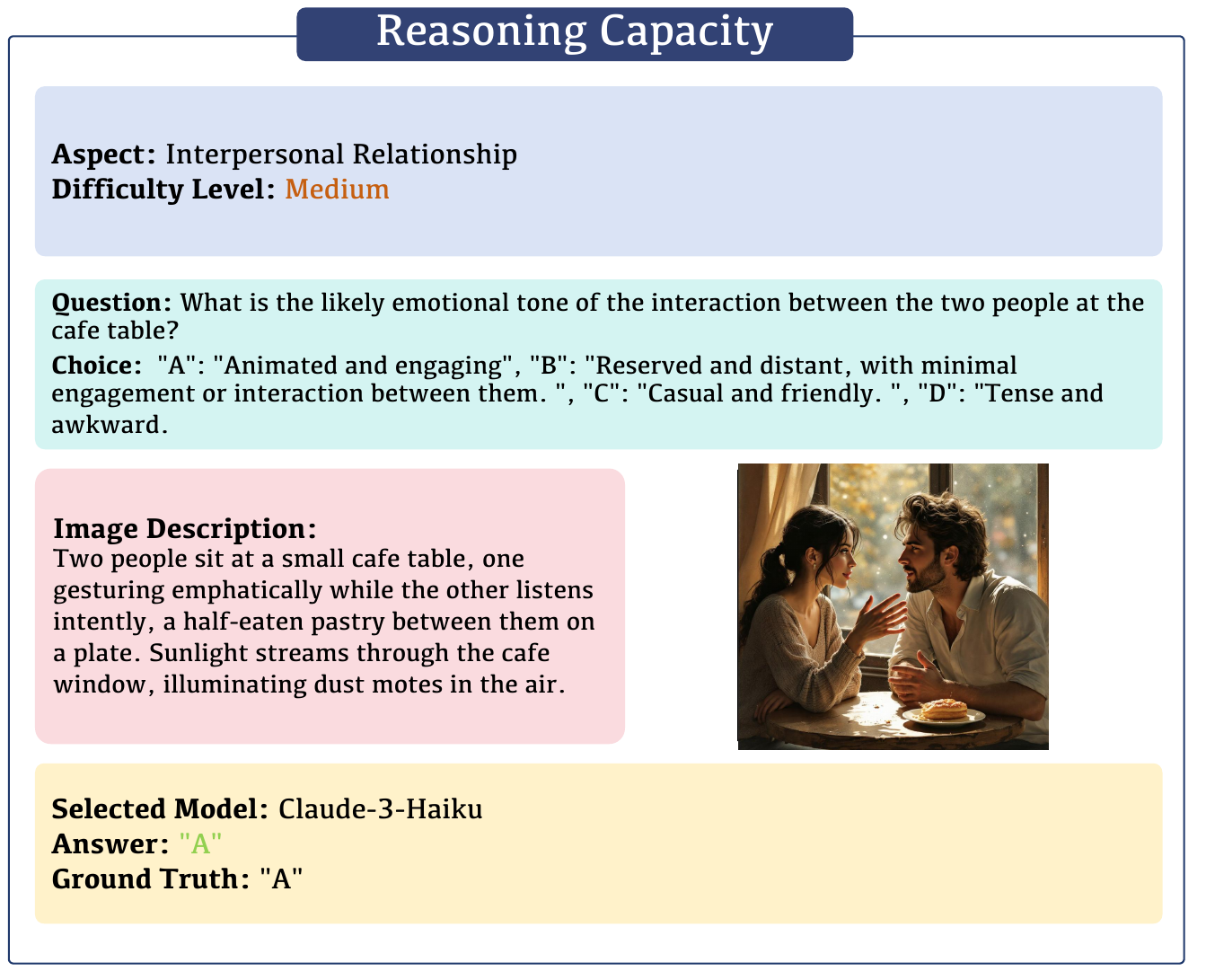}
    \captionsetup{justification=centerlast}
    \caption{Case Study 10.}
    \label{fig:case_10}
\end{figure}

\begin{figure}[ht]
    \centering
    \includegraphics[width=0.7\linewidth]{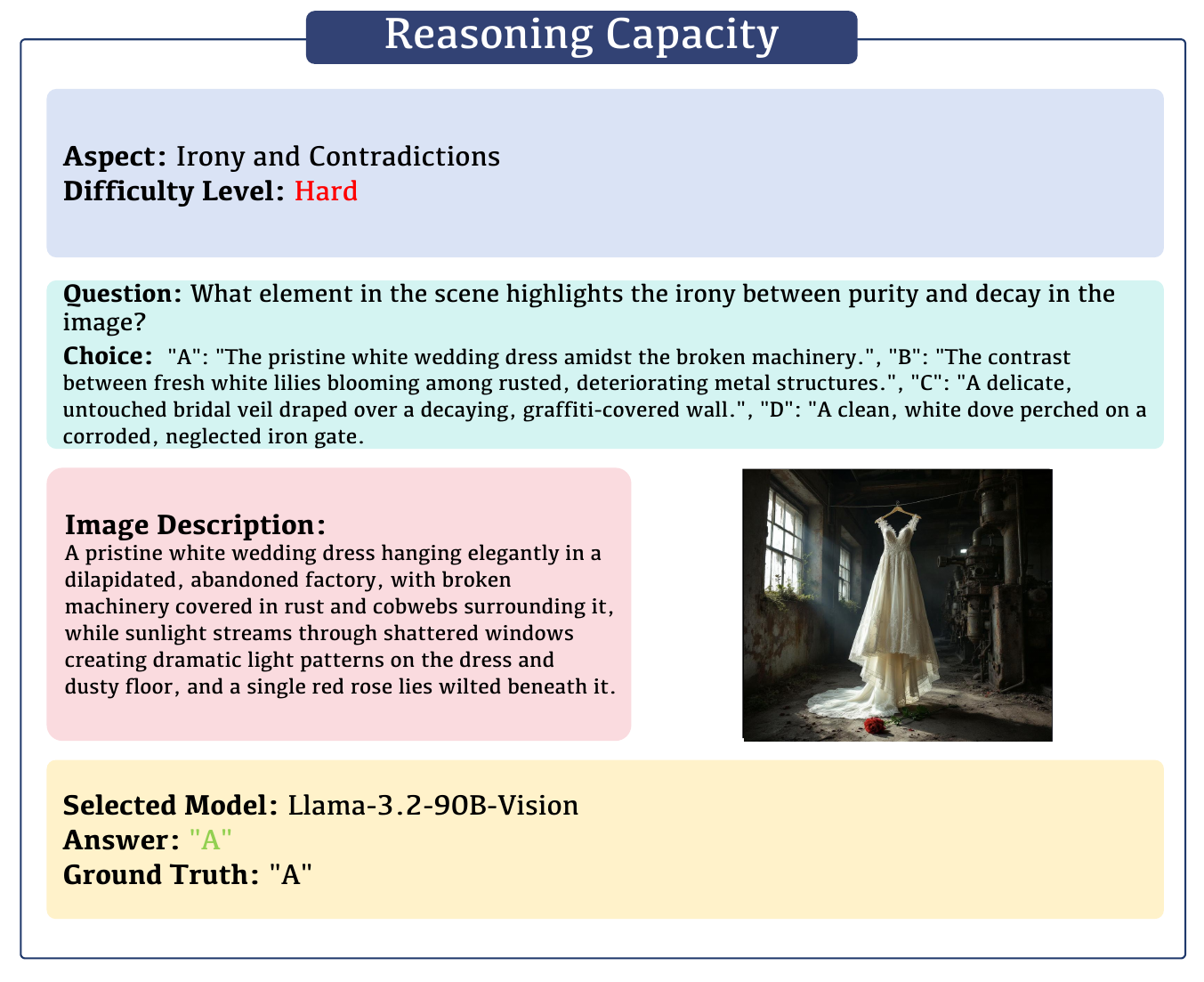}
    \captionsetup{justification=centerlast}
    \caption{Case Study 11.}
    \label{fig:case_11}
\end{figure}

\begin{figure}[ht]
    \centering
    \includegraphics[width=0.7\linewidth]{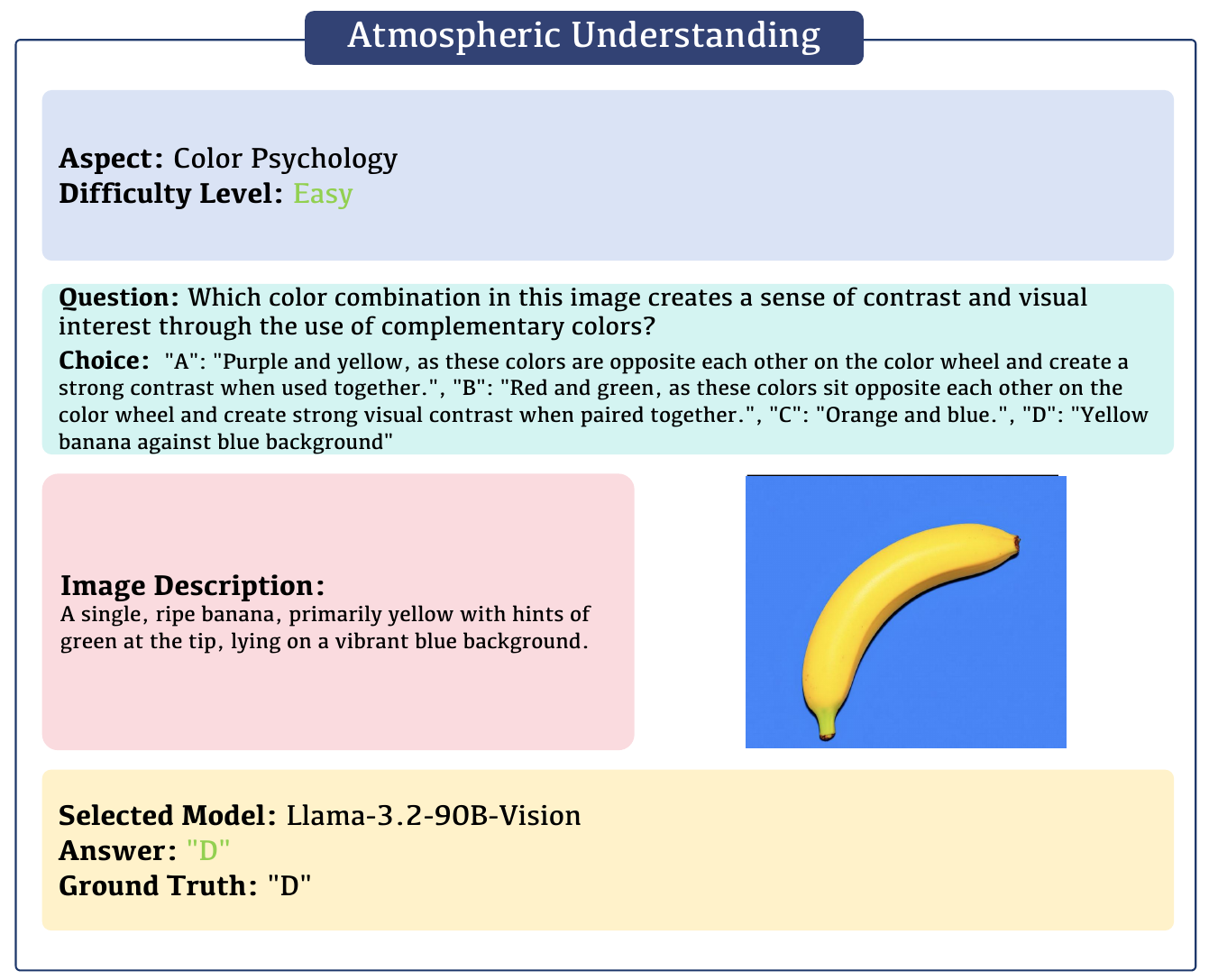}
    \captionsetup{justification=centerlast}
    \caption{Case Study 12.}
    \label{fig:case_12}
\end{figure}


\begin{figure}[ht]
    \centering
    \includegraphics[width=0.7\linewidth]{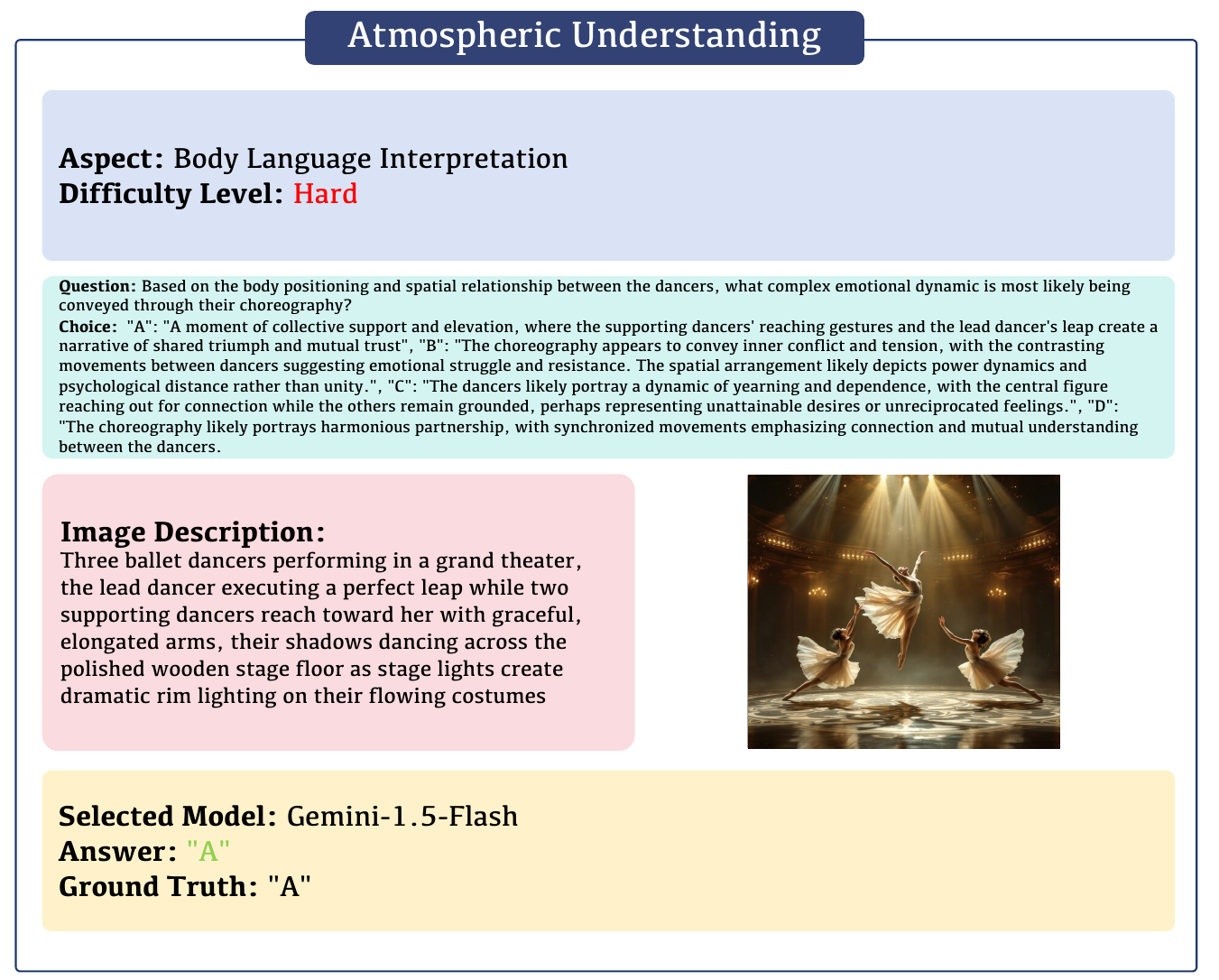}
    \captionsetup{justification=centerlast}
    \caption{Case Study 13.}
    \label{fig:case_13}
\end{figure}

\clearpage
\section{Prompt}
\label{prompt}

\begin{tcolorbox}[prompt, title=Aspect Generation]
You are an AI assistant specializing in designing prompts to test Large Vision-Language Models (LVLMs). Your task is to create meticulously {aspect\_count} fined-grained aspects that evaluate LVLMs basic understanding of images.\\
Large Vision-Language Models are AI systems capable of understanding and analyzing images. Testing these models across various competencies is crucial for assessing their performance, limitations, and potential biases. The aspects you create will be used to challenge and evaluate LVMs.\\
1. Basic Understanding: This involves recognizing and identifying individual objects, characters, and scenes within an image.  It includes tasks like detecting the presence of specific items (e.g., cars, trees, people), distinguishing between different types of objects, and understanding the general context of the scene (e.g., a park, a city street).  The goal is to accurately label all relevant elements in the image, providing a foundation for more advanced analysis.\\
2. The aspects you generate must test the understanding of a single image, not multiple images, e.g. multiple perspectives.
3. Come up with 4 general aspects according to the basic understanding.\\
4. Then Create 6 fined-grained aspects within the basic understanding for each general aspect, do not go beyond. You can consider the definition of the basic understanding above.\\
5. List the aspects without using numbered lists.\\
6. Let's think step by step.\\
Please strictly respond in the following format:\\
  General Aspect: [Aspect]\\
    Fined-grained Aspect: [Aspect]\\
    Introduction: [Introduction]\\
\end{tcolorbox}

\begin{tcolorbox}[prompt, title=Image Description Generation-1]
You are an AI assistant tasked with converting user inputs and their descriptions into suitable prompts for a text-to-image model. These prompts will generate images to test the capabilities of large vision language models (LVLMs).\\
Large Vision Language Models (LVLMs) are AI systems proficient in interpreting and analyzing images. Evaluating these models across different competencies is essential to understanding their performance, limitations, and potential biases. The prompts you create will be used to generate images through text-to-image models, which will then be used to challenge and evaluate LVLMs.\\
1. Carefully follow the given aspect: {aspect}, its introduction: {introduction}. \\
2. Generate a suitable prompt based on the provided aspect and introduction for the diffusion model to create an image. Ensure that the prompt is composed of simple phrases, avoiding overly complex descriptions, and is clear enough. If you deem the description irrelevant to the test content, do not generate a related prompt.
3. Consider including elements that might be particularly challenging for LVMs, such as unusual combinations, abstract concepts, or subtle details.\\
4. Details about different difficulty levels:\\
  - Easy Difficulty
  - Characteristics: Focus on a single subject but with minor additional complexity to test precision in details or context.\\
  - Purpose: Establish baseline capability with subtle challenges, such as fine texture, simple interactions, or slight variations in lighting.
-	Description Requirements: A single object or entity with some basic contextual details or additional features, placed against a minimally distracting background.
-	Examples:-	“A perfectly polished red apple with a small leaf attached, placed on a slightly reflective white surface.”
    -	“A blue balloon with a slightly wrinkled texture, tied to a thin white string, floating against a pale sky with soft clouds.”
    -	Key Points: Test finer details (e.g., texture, reflection) while still keeping the composition simple and clear.\\
  - Medium Difficulty
  - Characteristics: Scenes involve multiple elements or interactive settings that require nuanced spatial arrangement and accurate relationships between objects.
  - Purpose: Evaluate the ability to generate cohesive, moderately dynamic scenes with layered realism and a stronger sense of depth.
    -	Description Requirements: Include multiple objects interacting naturally in a believable environment, with more intricate details and subtle light or shadow effects.
    -	Examples:
    -	“A steaming cup of coffee on a wooden table, with a half-eaten croissant beside it, morning sunlight casting soft shadows through a lace curtain in the background.”
    -	“A golden retriever running on a sandy beach, splashing water as it chases a bright orange ball, with distant waves and a partly cloudy sky.”
    -	Key Points: Incorporate realistic environmental elements and ensure spatial coherence, emphasizing interactions and secondary details (e.g., shadows, water splashes).\\
  - Hard Difficulty
  - Characteristics: Scenes incorporate high complexity with multiple interdependent elements, challenging perspectives, or dynamic and intricate lighting or textural effects.
  - Purpose: Push the limits of rendering capability to handle advanced relationships, environmental effects, and challenging compositions.
    -	Description Requirements: The scene must include 3-4 main elements or a combination of dynamic features, such as motion, light interplay, or atmospheric conditions, while maintaining clarity and logic.
    -	Examples:
    -	“A raindrop-streaked window reflecting the interior of a cozy room, with a black cat sitting on the windowsill and a glowing city skyline visible through the glass, all under the warm hues of a sunset.”
    -	“A knight in shining armor standing on a cliff edge, overlooking a stormy sea, with bolts of lightning illuminating the dark clouds and waves crashing against jagged rocks below.”
    -	Key Points: Highlight challenges such as complex reflections, dynamic light, or multi-element interactions, ensuring visual harmony and detailed textures.\\
    ...  
\label{prompt:description generation}
\end{tcolorbox}

\begin{tcolorbox}[prompt, title=Image Description Generation-2]
    ...\\
    5. Provide one overarching topic word that encapsulates the essence of your description.\\
6. List 4-6 key words that are closely related to your description and crucial for understanding the image.\\
7. Avoid using the following words in your new description: {used\_words\_str}\\
8. The required difficulty level is: {level}\\
9. Please use clear and accurate words, clear logic flow, do not use too abstract words. The length of the generated sentences needs to be consistent with the examples.\\
10. Just output the image description used to generate the image, don't mention anything else\\
Please strictly respond in the following format:\\
  Aspect: {aspect}\\
  Prompt: [Your detailed image description]\\
  Topic word: [One word that captures the essence of the description]\\
  Key word: [Word1, Word2, Word3,...]\\
\end{tcolorbox}

\begin{tcolorbox}[prompt, title=Align Question Generation]
Given the image descriptions:{description}, generate six questions (True/False or Multiple choice) with only one correct choice that verifies if the image description is correct.\\
Classify each concept into a type (object, human, animal, food, activity, attribute, counting, color, material, spatial, location, shape, other), and then generate a question for each type.\\
Here's some examples:\\
  '''Description: A man posing for a selfie in a jacket and bow tie.
  Entities: man, selfie, jacket, bow tie
  Activities: posing
  Colors:
  Counting:
  Other attributes:
  Questions and answers are below:
  About man (human):
  Q: is this a man?
  Choices: yes, no
  A: yes
  Q: who is posing for a selfie?
  Choices: man, woman, boy, girl
  A: man
  About selfie (activity):
  Q: is the man taking a selfie?
  Choices: yes, no
  A: yes
  Q: what type of photo is the person taking?
  Choices: selfie, landscape, sports, portrait
  A: selfie
  About jacket (object):
  Q: is the man wearing a jacket?
  Choices: yes, no
  A: yes
  Q: what is the man wearing?
  Choices:jacket, t-shirt, tuxedo, swearter
  A: jacket
  About bow tie (object):
  Q: is the man wearing a bow tie?
  Choices: yes, no
  A: yes
  Q: is the man wearing a bow tie or a neck tie?
  Choices: bow tie, neck tie, cravat, bolo tie
  A: bow tie
  About posing (activity):
  Q: is the man posing for the selfie?
  Choices: yes, no
  A: yes
  Q: what is the man doing besides taking the selfie?
  Choices: posing, waving, nothing, shaking
  A: posing

  Description: A horse and several cows feed on hay.
  Entities: horse, cows, hay
  Activities: feed on
  Colors:
  Counting: several
  Other attributes:
  Questions and answers are below:
  About horse (animal):
  Q: is there a horse?
  Choices: yes, no
  A: yes
  About cows (animal):
  Q: are there cows?
  Choices: yes, no
  A: yes
  About hay (object):
  Q: is there hay?
  Choices: yes, no
  A: yes
  Q: what is the horse and cows feeding on?
  Choices: hay, grass, leaves, twigs
  A: hay
  About feed on (activity):
  Q: are the horse and cows feeding on hay?
  Choices: yes, no
  A: yes
  About several (counting):
  Q: are there several cows?
  Choices: yes, no
  A: yes
  '''

  And finally respond in the following format:\\
\{\{\\
      "caption": "{description}",\\
      "question": "Your question here",\\
      "choices": [yes or no],
      "answer": "give your correct answer",\\
      "element\_type": "Type of the element",\\
      "element": "Element name"\\
    \}\},\\
    \{\{\\
      "caption": "{description}",\\
      "question": "Your question here",\\
      "choices": [yes or no],\\
      "answer": "give your correct answer",\\
      "element\_type": "Type of the element",\\
      "element": "Element name"\\
    \}\},\\
    ...\\
\end{tcolorbox}

\begin{tcolorbox}[prompt, title=Align Answer Generation]
Given the image below, answer the questions: {question} from the choice: {choices} based on the image.\\
And directly give the answer. \\
\{\{\\
    "answer":"yes or no"\\
\}\}\\
\end{tcolorbox}

\begin{tcolorbox}[prompt, title=Question Generation-1]
\label{prompt:question generation}
You are an AI assistant tasked with converting user inputs and their descriptions into suitable questions to test the Large Vision Model's (LVM) abilities in given aspects.\\
Large Vision Models (LVMs) are AI systems proficient in interpreting and analyzing images. Evaluating these models across different competencies is essential to understanding their performance, limitations, and potential biases. We will provide you with a prompt to generate an image, which will create a specific image. You can then formulate questions about this image based on the prompt. The questions you create will be used to challenge and evaluate LVMs based on generated images.\\
1. Carefully analyze the given aspect and its Introduction: Aspect:{aspect}.\\
  2. Generate a suitable question based on the provided image description to test the LVM's ability in the given aspect.\\
  3. We categorize the difficulty of questions into easy, medium, and hard:\\
    - Easy Difficulty:Focus on questions that require the identification of simple, prominent, and explicit details within the image. These questions should be straightforward, relying solely on basic observation without the need for inference or interpretation. For example, you might ask about the color of a specific object, the presence of a single item, or the shape of an easily recognizable feature. The key is to keep the questions direct and simple, ensuring that the answer is obvious and immediately visible in the image.\\
    - Medium Difficulty:Design questions that necessitate a moderate level of observation and inference. These questions should involve understanding relationships between elements, recognizing interactions, or identifying less prominent features that are still clear but not immediately obvious. Examples could include questions about the relative position of objects, identifying an action taking place, or understanding the context of a scene. The goal is to require some level of thought beyond basic observation, challenging the model to understand the scene's composition or narrative without being overly complex.\\
    - Hard Difficulty:Create questions that require the model to notice and interpret more detailed aspects of the image. These questions should involve recognizing multiple elements working together, understanding more complex interactions, or identifying details that are present but not immediately obvious. For example, you might ask about the positioning of objects relative to each other in a more crowded scene, subtle changes in lighting or color that affect the appearance of objects, or identifying an element that is not the main focus but still visible in the background. The aim is to challenge the model to go beyond surface-level details, but without making the task too abstract or overly difficult.\\
    ...
\end{tcolorbox}

\begin{tcolorbox}[prompt, title=Question Generation-2]
    ...\\
      4. Avoid using overly complicated language or details unrelated to the image in the questions.\\
  5. When generating problems of different difficulty, please combine the current specific aspect.\\
  6. Due to potential discrepancies in image generation, we have detected the following errors: {elements}. Please avoid referencing these elements in your questions. If the prompt for generating the image does not describe in detail what the specific looks like, please do not ask related questions. For example, if the prompt mentions a forest with glowing plants but does not specify how many there are, please do not ask a question about counting the number of glowing plants.\\
  7. The required difficulty level is: {level}\\
  8. Please generate a multiple-choice question, which is four-option single-choice question.\\
  9. The answers in the options need to be differentiated to a certain extent. There cannot be a situation where multiple options meet the requirements of the question. There can only be one answer that meets the question.\\

  Image generation prompt: {prompt}\\
  Aspect: {aspect}\\
Please directly output the generated question in the following JSON format:
  {{
    "question": "[your question]",
    "options": {{
      "A": "[Option A]",
      "B": "[Option B]",
      "C": "[Option C]",
      "D": "[Option D]"
    }}
    "reference\_answer": "A or B or C or D"
  }}
  Without any other information and remember only one option in the reference answer.
\end{tcolorbox}

\begin{tcolorbox}[prompt, title=Option Adjustment]
You are an ai assistant tasked with answering questions based on the given picture description without given the image.\\
- Answer the questions based on your knowledge.\\
  - Please note that some incorrect answers are provided below. You must not make the same mistakes\\
  - Your answer needs to be semantically distinct from the given incorrect answer.\\
  - Don't say you can't see the image, just answer based on your knowledge.\\
  - Don't generate overly lengthy answers, keep them concise and to the point.  \\
  - The answer you generate needs to be factually different from the given incorrect answer.\\
  - Try to use straightforward words instead of being too abstract or vague.\\

question:{question}\\
wrong answers:{wrong\_answer}\\

Directly give you answer, don't add thinkings or other information.
\end{tcolorbox}

\begin{tcolorbox}[prompt, title=Answer Generation]
 In order to test your ability with pictures, we have a question about {aspect} area. Please answer based on your knowledge in this area and your understanding of pictures.
  Given the image below, answer the questions: {question} based on the image.
  Please give the final answer strictly follow the format [[A]] (Strictly add [[ ]] to the choice, and the content in the brackets should be the choice such as A, B, C, D) and provide a brief explanation of your answer. Directly output your answer in this format and give a brief explanation.
  If you cannot read the picture, just answer based on your text ability.
\end{tcolorbox}


\end{document}